\documentclass[final,3p,times,authoryear]{elsarticle}
    \usepackage[table]{xcolor}

\usepackage{natbib}
\renewcommand\harvardurl[1]{\textbf{URL:} \url{#1}}
\usepackage{hyperref}

\usepackage{amssymb}
\usepackage{lipsum}

\usepackage{amsmath}
\usepackage{lscape} 
\usepackage{booktabs}
\usepackage{array}
\usepackage{caption}
\usepackage{subcaption}
\usepackage{multirow}
\usepackage{adjustbox}
\usepackage{amssymb}
\usepackage{textgreek}
\usepackage{tcolorbox}
\usepackage{xcolor}
\usepackage{makecell}
\usepackage{tabularx}
\usepackage{makecell}  
\usepackage{pifont} 
\usepackage{pifont} 

\newcommand{\cmark}{\ding{51}} 

\usepackage{threeparttable} 

\usepackage{algorithm, algorithmic}
\newcolumntype{F}[1]{%
    >{\raggedright\arraybackslash\hspace{0pt}}p{#1}}%
\newcolumntype{T}[1]{%
    >{\centering\arraybackslash\hspace{0pt}}p{#1}}%

\hypersetup{urlcolor=blue, colorlinks=true} 

\usepackage{graphicx} 
\usepackage{tabularx}
\usepackage{booktabs}
\usepackage{makecell} 

\newcolumntype{C}{>{\centering\arraybackslash}X}

\journal{Transportation Research: Part C}

\begin{document}

\begin{frontmatter}




\title{T-STAR: A Context-Aware Transformer Framework for Short-Term Probabilistic Demand Forecasting in Dock-Based Shared Micro-Mobility}


\author[first]{Jingyi Cheng}
\author[first]{Gonçalo Homem de Almeida Correia}
\author[first]{Oded Cats}
\author[first]{Shadi Sharif Azadeh}
\affiliation[first]{organization={Transport and Planning, Delft University of Technology}, 
            city={Delft},
            country={Netherlands}}

\begin{abstract}
Reliable short-term demand forecasting is essential for managing shared micro-mobility services and ensuring responsive, user-centered operations. This study introduces T-STAR (Two-stage Spatial and Temporal Adaptive contextual Representation), a novel transformer-based probabilistic framework designed to forecast station-level bike-sharing demand at a 15-minute resolution. T-STAR addresses key challenges in high-resolution forecasting by disentangling consistent demand patterns from short-term fluctuations through a hierarchical two-stage structure. The first stage captures coarse-grained hourly demand patterns, while the second stage improves prediction accuracy by incorporating high-frequency, localized inputs, including recent fluctuations and real-time demand variations in connected metro services, to account for temporal shifts in short-term demand. Time series transformer models are employed in both stages to generate probabilistic predictions.
Extensive experiments using Washington D.C.’s Capital Bikeshare data demonstrate that T-STAR outperforms existing methods in both deterministic and probabilistic accuracy. The model exhibits strong spatial and temporal robustness across stations and time periods. A zero-shot forecasting experiment further highlights T-STAR’s ability to transfer to previously unseen service areas without retraining. These results underscore the framework’s potential to deliver granular, reliable, and uncertainty-aware short-term demand forecasts, which enable seamless integration to support multimodal trip planning for travelers and enhance real-time operations in shared micro-mobility services.

\end{abstract}




\end{frontmatter}




\section{Introduction}
\label{introduction}

Shared micro-mobility services have become integral to urban transportation networks, offering flexible, low-emission options for short-distance and first- and last-mile travel \citep{yang2019spatiotemporal}. 
Among them, bike-sharing services have seen rapid global adoption, with large-scale networks deployed across major cities to enhance accessibility \citep{ma2020comparison}. 
As a key component of Mobility-as-a-Service (MaaS) ecosystems, bike-sharing services interact closely with public transport as well as other on-demand modes, facilitating seamless multimodal journeys within complex urban environments \citep{cleophas2019collaborative}. 
However, their operational efficiency is often constrained by spatio–temporal imbalances between supply and demand, leading to an inefficient resource allocation and service disruptions. 
Accurate short-term demand forecasting is therefore essential for supporting real-time operational decisions, including fleet rebalancing and multimodal trip recommendations \citep{zhao2025research, chiariotti2020bike, zhang2025stochastic}.
With shared (micro-)mobility becoming increasingly integrated into multimodal transport systems, predictive tools that can handle these complexities are essential to support reliable, efficient, and user-centered operations.

Despite encouraging progress in short-term demand forecasting for bike-sharing systems, several important challenges remain that limit their operational effectiveness. 
Demand is shaped by complex, interrelated factors, ranging from exogenous influences like weather variability, public transport service disruptions, and other sudden local events to endogenous constraints such as fleet availability and inter-station demand interactions \citep{wilkesmann2023determinants, jia2021exploring,huang2021spatiotemporal}. Although some studies have started incorporating multi-source contextual data \citep{22_xu2023multi}, integrating these dynamic signals in a structured and interpretable way remains technically demanding.
Moreover, shared micro-mobility studies emphasize that high-resolution forecasts, such as at the station level and in short intervals, are important for effective operations\citep{chiariotti2020bike}.
However, such fine-grained predictions introduce modeling challenges due to data sparsity, intermittency, and noise, particularly if they are to be performed at intervals of 15 minutes or less \citep{14_kourentzes2013intermittent}. Additionally, while probabilistic forecasting offers important advantages for uncertainty quantification and risk-aware planning \citep{zhao2025research}, 
related efforts in the forecasting literature remains limited.
Despite growing demand for such capabilities, it remains largely unknown how to generate high-resolution, uncertainty-aware forecasts that effectively integrate heterogeneous contextual signals while maintaining robustness to sparsity and noise \citep{xu2018station, 13_gammelli2022predictive}. 
These gaps underscore the need for context-aware probabilistic forecasting frameworks that can support effective real-time decision-making related to urban shared micro-mobility services.

To address the challenges of fine-grained, uncertainty-informed demand forecasting in dock-based bike-sharing systems, this study proposes the Two-stage Spatial and Temporal Adaptive contextual Representation (T-STAR) framework. While prior research has explored station-level and probabilistic forecasting, existing methods often struggle to balance the inherent noise and sparsity typical of ultra-short-term intervals \citep{liang2023cross, 22_xu2023multi, 13_gammelli2022predictive}. Unlike single-stage models that may conflate stochastic noise with recurring patterns, T-STAR distinguishes itself by employing a hierarchical architecture designed to explicitly navigate this resolution-uncertainty trade-off. Built on a Transformer backbone, the framework acts as a global predictor that generalizes across stations to support scalable, real-time forecasting. In the first stage, the model captures broad demand expectations using hourly aggregated network features; the second stage then refines these representations with fine-grained, station-specific contextual signals to model high-frequency fluctuations. This hierarchical design allows T-STAR to disentangle recurrent temporal patterns from localized disruptions, providing robust, uncertainty-aware predictions specifically optimized for the 15-minute resolution required for proactive rebalancing operations.

We demonstrate and validate the Transformer-based T-STAR framework through 15-minute, station-level demand forecasting experiments using real-world data from Washington D.C.’s Capital Bikeshare system. The evaluation covers both deterministic and probabilistic forecasting accuracy. We analyze spatial robustness by comparing model performance across stations with varying demand profiles, and assess temporal stability by examining prediction quality at each time step, including intervals under extreme conditions and abnormal observations. Additionally, a zero-shot forecasting experiment on an unseen service area is conducted to test model’s adaptability and generalization capability. Collectively, the results demonstrate that T-STAR delivers accurate and robust predictions, making it a reliable and scalable tool for large-scale urban shared micro-mobility systems. A coding and application demo of the Transformer-based T-STAR model is provided through a public GitHub repository associated with this study\footnote{\url{https://github.com/RinaPiggy/T-STAR}}.

The remainder of this study is organized as follows. Section \ref{chapter:literature} reviews the related literature and identifies the gaps. 
Section \ref{chapter:problem_formulation} introduces the formulation of probabilistic short-term demand forecasting problem for a dock-based bike-sharing system. We then explain
the proposed T-STAR framework in Section \ref{chapter:methods}, detailing its hierarchical architecture and contextual embedding strategy. Section \ref{chapter:case} describes the case study, evaluation metrics, and design of benchmarking experiments. In section \ref{chapter:results_and_discussion} we analyze the results of the forecasting experiments and discuss the key findings. Finally, Section \ref{chapter:conclusion} concludes this study and suggests directions for future research.


\section{Literature Review}
\label{chapter:literature}

\subsection{Demand Forecasting Approaches for One-Way Bike-Sharing Systems}

State-space models (SSMs) have long served as the foundation for classical time series forecasting, offering a parametric framework that decomposes demand series into components such as trend, seasonality, and autoregressive error terms \citep{2_durbin2012time}. Among these, Seasonal Auto-Regressive Integrated Moving Average (SARIMA) models have been frequently applied to forecast bike-sharing demand \citep{3_yoon2012cityride, xu2018station}. Despite their interpretability, SSMs are ill-suited for short-term bike-sharing scenarios, where demand is highly variable, zero-inflated, and non-negative \citep{5_lim2022probabilistic}. Furthermore, the need to model each station’s dynamic demand time series individually poses scalability challenges for large-scale networks, making SSMs computationally expensive to be adapted for urban bike-sharing systems \citep{5_lim2022probabilistic}.

To capitalize on the availability of large-scale bike-sharing demand data, recent studies have increasingly adopted machine learning (ML) approaches for forecasting. Compared to traditional state-space models, ML methods can efficiently process high-dimensional inputs and incorporate diverse contextual features to capture the complex, dynamic patterns of urban micro-mobility demand. Non-parametric tree-based models, such as Random Forest (RF) and eXtreme Gradient Boosting (XGBoost), have been widely applied in this context \citep{6_gao2022using,7_sathishkumar2020using,8_leem2024enhancing}. Using data from the Seoul bike-sharing system, \cite{6_gao2022using} and \cite{7_sathishkumar2020using} demonstrated that tree-based regressors can generate accurate city-level forecasts by leveraging multiple external inputs. 
Deep learning has shown promising performance in forecasting intermittent and highly variable demand series \citep{14_kourentzes2013intermittent}. To better capture the complex temporal and spatial dynamics in bike-sharing networks, recent studies have increasingly focused on neural networks for demand forecasting \citep{9_zhang2018short, xu2018station, 5_lim2022probabilistic, 13_gammelli2022predictive, 15_lin2018predicting, li2023improving}. Temporal dependencies have been modeled using recurrent architectures, such as Recurrent Neural Networks (RNNs) and long short-term memory networks (LSTMs). \cite{zhang2018short} designed an LSTM-based predictor to learn long-term pattern while adaptively accounting for short-term fluctuations. Similarly, \cite{xu2018station} applied LSTMs for short-term inflow and outflow prediction in a large-scale free-floating system. To exploit shared spatial patterns across the network and model spatial dependencies among stations or service zones, studies have looked into convolutional neural networks (CNNs), graph neural networks (GNNs), and graph convolutional networks (GCNs). For example, \cite{15_lin2018predicting} proposed a GCN-based predictor that uncovers heterogeneous pairwise correlations between stations. \cite{li2023improving} proposed a hybrid CNN-LSTM model to jointly capture spatiotemporal demand patterns.

Transformer models, originally developed for natural language processing, have demonstrated promising performance in time series forecasting, proving particularly effective at capturing the spatiotemporal complexity characteristic of large-scale shared mobility systems \citep{zerveas2021transformer, 17_hu2023high, 18_li2024demand}.
Unlike the sequential structure in RNNs, Transformer models leverage self-attention to capture both short- and long-range temporal dependencies, enabling more flexible and efficient learning of complex seasonality and non-linear demand patterns across stations or service zones \citep{17_hu2023high, 18_li2024demand}.
Their parallel processing structure facilitates the seamless integration of diverse multivariate inputs, such as weather, time, and socioeconomic indicators. Models like the Interpretable Hierarchical Transformer (IHTF) \citep{17_hu2023high} and the Temporal Fusion Transformer (TFT) \citep{lim2021temporal} enhance this capacity through variable selection networks and multi-head attention mechanisms that dynamically identify the most relevant features at each time step. Acting as global predictors, these Transformer models learn shared temporal-spatial representations across the network while forecasting adaptively to the unique dynamics of individual locations. 

Transformers offer, next to their modeling capacity, several practical advantages over traditional sequential models. The self-attention architecture enables full parallelization of historical data processing, leading to significantly faster training on large datasets compared to RNNs or LSTMs, which rely on sequential computation \citep{vaswani2017attention}. Transformers also demonstrate strong generalizability and have been successfully applied in cross-location and zero-shot forecasting tasks in mobility contexts, making them suitable for dynamic systems where new service areas are frequently introduced \citep{19_solatorio2023geoformer}. Moreover, Transformer supports continuous fine-tuning from pre-trained models, enabling rapid adaptation to evolving urban patterns with limited new data \citep{19_solatorio2023geoformer}. These scalability, adaptability, and transferability properties make Transformers a promising forecasting solution for data-rich, yet rapidly changing shared micro-mobility services.

\textcolor{black}{While most past studies on shared micro-mobility have focused on point forecasting methods, recent research has increasingly emphasized the importance of probabilistic forecasting. Accurately probabilistic demand forecasting is critical for risk-aware management, as it quantifies the uncertainty inherent in fluctuating demand rather than providing single-point estimates. As reviewed by \cite{cheng2024recent}, these models equip managers to mitigate potential adverse conditions by evaluating the full distribution of possible outcomes \citep{zhao2025research, 22_xu2023multi, 13_gammelli2022predictive}. Urban modes like ride-hailing, public transit, and bike-sharing share foundations in spatiotemporal dependence and heterogeneous context. 
Related work addressing demand uncertainty across various mobility sectors can be broadly categorized into three methodological frameworks: quantile-based estimation, generative and latent-variable modeling, and parametric density estimation.}

\textcolor{black}{Quantile-based methods bypass the need for explicit distributional assumptions by directly estimating empirical quantiles or prediction intervals. These approaches are particularly valued for their robustness against outliers and non-standard noise profiles. For instance, Quantile Regression Forests (QRF) and Temporal Fusion Transformers (TFT) have been successfully deployed to estimate demand quantiles for both on-demand ride-hailing and scheduled transit services \citep{23_lei2014distribution, 24_cheng2025short, wang2025co}. While these models provide flexible interval estimates, they often lack a closed-form probability density function, which can complicate their integration into certain stochastic optimization pipelines.}

\textcolor{black}{To capture more complex, high-dimensional uncertainty, recent research has pivoted toward Generative Models and Neural Processes (NPs). These frameworks typically learn a mapping from input features to a latent representation $z$, which represents the underlying stochasticity of the demand process. An empirical distribution of future demand by performing repeated sampling of $z$. Notable applications include VAE-based STGCNs for ride-hailing \citep{peng2025uncertainty} and Bidirectional Spatial-Temporal Transformer NPs for ride-sourcing \citep{li2024demand}. Furthermore, \cite{22_xu2023multi} extended this logic to dock-based bike-sharing by integrating Transformer encoders to refine spatiotemporal feature extraction, while \cite{13_gammelli2022predictive} introduced a Variational Poisson RNN to estimate posterior rates for bike-sharing pickup and drop-off activities. However, latent compression mechanisms of generative modeling often dismiss high-frequency volatility as noise through posterior collapse \citep{dai2020usual, li2022generative}. Consequently, prioritizing global distribution over sharpness produces conservative forecasts that cannot accurately capture the precise transitions required for high-resolution, sparse demand patterns \citep{choi2022hierarchical}.}

\textcolor{black}{In contrast, Parametric Density Estimation methods assume that demand follows a predefined probability distribution and utilize deep learning to estimate its parameters. Although this approach carries the risk of distributional misspecification, it provides a mathematically grounded framework that is highly tractable for downstream stochastic optimization problems. Based on a Normal distribution assumption, \cite{li2020graph} utilized GNNs for robust public transit forecasting. Prob-GNN was proposed by \cite{wang2024uncertainty} to model multi-modal demand via truncated Gaussian distributions. \cite{5_lim2022probabilistic} investigated the impact of different stochastic priors by fitting normal, truncated normal, and negative binomial distributions with DeepAR to generate probabilistic bike-sharing demand forecasts.}

\subsection{The Granularity of Demand Forecasting Problems in Shared Micro-Mobility}

Demand forecasting in shared micro-mobility systems can be categorized based on its spatial and temporal granularity, which are suitable to assist different operations. Coarse-grained forecasts, such as city-wide daily demand predictions, are suitable for long-term and strategic planning decisions. In contrast, spatially and temporally fine-grained forecasts, such as station-level per 15-minute forecasting, are essential for operational decisions that require both timely and accurate predictions. 


The literature on demand forecasting for bike-sharing services can be categorized by the spatial granularity of the prediction task: city-level, zonal or cluster-level, and station-level forecasting.
Early studies addressed this problem at the city level, predicting the total number of rental requests across the entire network for future time periods \citep{ giot2014predicting}. To better capture spatial heterogeneity in usage patterns, subsequent work shifted toward regional-level forecasting. For example, some studies applied clustering algorithms such as K-means to group stations together based on similarity, then forecasts aggregate demand for each cluster \citep{li2015traffic}. Building on this, \cite{chen2016dynamic} proposed dynamic clustering frameworks that incorporate real-time contextual signals, such as weather or events, to adaptively redefine spatial groupings over time, improving the model’s responsiveness to changes in demand distribution. These cluster-then-predict approaches aim to enhance forecast accuracy by aggregating sparse station-level signals within semantically or contextually coherent regions, thereby reducing noise and data sparsity. 
To overcome the loss of local detail inherent in cluster-based methods \citep{feng2024adaptive}, recent studies have focused on station-wise demand forecasting, leveraging high-resolution data and advanced modeling techniques to provide more precise and actionable insights \citep{22_xu2023multi, 5_lim2022probabilistic, 13_gammelli2022predictive, feng2024adaptive}.

In short-term demand forecasting for dock-based bike-sharing services, temporal granularity plays a crucial role in determining the operational applicability of predictive insights. Most studies adopt an hourly aggregation level to balance modeling complexity and practical relevance \citep{22_xu2023multi, li2023improving, 15_lin2018predicting}. Some researchers have explored coarser intervals, such as splitting the day into peak and off-peak periods or forecasting in 4-hour blocks, to capture broader demand trends \citep{chen2016dynamic, liang2023cross}. However, for operations with short lead times, such as real-time and continuous rebalancing \citep{chiariotti2020bike}, higher temporal resolution is essential. To this end, several studies have investigated ultra-short-term forecasting at 15- or 20-minute intervals \citep{xu2018station, 13_gammelli2022predictive, ma2022short, feng2024adaptive}, enabling a more proactive and precise management of bike-sharing systems. Focusing on a zone-level forecasting problem, \cite{xu2018station} analyzed the tradeoff between temporal granularity and predictive accuracy by evaluating multiple interval lengths (10, 15, 20, and 30 minutes) across both state-space and neural network models. Their findings suggest that, while shorter intervals provide more timely information, longer intervals yield lower forecast error due to reduced noise and variability. 

Studies have shown that spatially and temporally fine-grained predictions are crucial for enabling real-time operations in bike-sharing systems, such as dynamic fleet rebalancing, where vehicle routes and allocations are continuously updated \citep{chiariotti2020bike}, and multimodal trip recommendations that account for anticipated bike availability \citep{zhang2025stochastic}. However, generating accurate fine-grained forecasts is inherently challenging. Station-level demand is often sparse and highly volatile, especially during off-peak hours, and historical records may include irregular shocks caused by supply constraints, such as empty or full docking stations \citep{chen2016dynamic, ashqar2022network, xu2018station}. Although high-resolution forecasts offer greater operational responsiveness, they tend to be more challenging due to noise and intermittency in demand data. In contrast, coarser temporal or spatial aggregation improves stability but sacrifices the detail needed for localized decision-making. To address this tradeoff, \cite{chen2020detecting} proposed a Bayesian hierarchical model that captures long-term demand shifts at the city level in the upper layer, while modeling daily station-level demand in the lower layer, incorporating dynamic features such as weather and holidays. Accounting for spatial dependencies across different levels of granularity, \cite{khalesian2024improving} proposed a hierarchical reconciliation framework that refines zone-level traffic demand forecasts using aggregated regional predictions.
Although the study of hierarchical modeling in bike-sharing demand forecasting remains relatively limited, this direction appears promising for integrating coarse- and fine-grained signals to enhance the accuracy and robustness of ultra-short-term, station-level predictions.


\subsection{Determinants of Bike-Sharing Demand}
\label{subsect:contextual_features}
Accurate short-term demand forecasting in bike-sharing systems relies not only on historical usage patterns but also on a variety of contextual factors that influence user demand. These factors capture both recurring patterns and sudden fluctuations, enabling models to effectively reflect the spatial, temporal, and environmental dynamics of the system. In this subsection, we review key contextual features discussed in the literature, including temporal patterns, fleet availability and station capacity, connectivity to public transit, weather conditions, and latent events.

Like other urban transit services, bike-sharing systems exhibit strong cyclic demand patterns that align with daily commuting routines, typically peaking during morning and evening rush hours \citep{zhao2019effect, faghih2014land, cats2022unravelling}. 
To capture these recurring trends, seasonal features, such as hour-of-the-day and day-of-the-week, are commonly included as fundamental inputs in short-term demand forecasting models.
However, these temporal patterns can vary significantly across stations due to local land use and neighborhood characteristics. For instance, stations at business and residential areas often experience demand peaks during commuting hours, while those located within pubs, restaurants, or entertainment districts tend to see higher demand in the evenings and on weekends 

Fleet availability and station capacity are key factors influencing short-term bike-sharing demand. High-capacity stations, typically located in areas with high passenger traffic, tend to attract greater usage due to the increased likelihood of bike availability and open docks for returns \citep{rixey2013station}. In contrast, low-capacity stations often experience suppressed rental activity, especially during peak periods, as limited bikes or docking spaces restrict access. These constraints give rise to censored demand, where true user intent is unobservable when stations are empty or full. As a result, discrepancies between actual and recorded demand can introduce bias and reduce forecasting accuracy. To mitigate this, recent studies have explicitly modeled the impact of censored demand, enhancing prediction accuracy and reliability in demand forecasting \citep{albinski2018performance, huang2021spatiotemporal}.

The interaction between bike-sharing and public transit has been widely studied within the broader context of Mobility on Demand (MoD) and Mobility as a Service (MaaS). This relationship can be either complementary or competitive. For example, passengers may shift away from public transit when station-based bike-sharing offers a faster, more affordable, or more direct alternative \citep{2_shaheen2015unraveling}. However, in lower-density or suburban areas, bike-sharing often plays a complementary role by bridging first- and last-mile gaps, thereby increasing the attractiveness of multimodal travel. Empirical studies have consistently shown that bike-sharing stations located near metro stations experience higher usage \citep{wang2021bikeshare, ma2019estimating, tarpin2020typology, tushar2024bikeshare}. 
There is a positive correlation between public transit passenger flows and nearby bike-sharing demand, especially during peak hours when commuters frequently combine multiple modes of transport.
During major transit disruptions, bike-sharing systems also help sustain network resilience by serving as alternative transport modes \citep{cheng2021role}. For example, during the 2016 WMATA metro closure, bike-sharing demand surged by 191\% in Washington D.C. \citep{jia2021exploring}. Reflecting this interdependence, recent forecasting models have begun incorporating public transit operations as exogenous features \citep{lv2021mobility, li2023improving}. These findings underscore the operational and predictive value of integrating multimodal connectivity into short-term demand forecasting frameworks.

Prior studies have shown that the demand for urban bike-sharing services is greatly influenced by external conditions \citep{zhao2019effect, wilkesmann2023determinants}.  Among these, weather conditions play a particularly significant role in shaping user behavior. For example, \cite{gebhart2014impact} found that cold temperatures, rainfall, and high humidity are associated with a decline in Capital Bikeshare trips. Conversely, warm, dry, and sunny days tend to correspond with increased demand, reflecting users’ greater willingness to travel by bike under favorable weather conditions \citep{gebhart2014impact, wang2021bikeshare}. 

Beyond commonly modeled factors, short-term bike-sharing demand can also be influenced by latent influences such as special events in the city, which are rare and has unclear impacts on the bike-sharing usage in the short run\citep{chen2016dynamic, jia2021exploring}. These events often occur irregularly and may not be directly captured by input features, yet they can induce sudden, localized shocks in demand. To account for such latent influences, recent studies have explored the use of neural networks to learn latent spatiotemporal representations from historical data \citep{li2019learning, ham2021spatiotemporal}. By leveraging deep learning models, these approaches infer the underlying patterns associated with unobserved factors, improving the ability of the predictors to generalize and respond to demand shifts that are not entirely explained by the observable variables.

\subsection{Summary of Literature and Research Gaps}

\begin{table}[h!]
\centering
\caption{Selected representative studies on short-term bike-sharing demand forecasting, categorized by spatial/temporal granularity, forecasting type, predictive model, and contextual inputs.}
\resizebox{\textwidth}{!}{%
\begin{tabular}{lccccc}
\hline
\textbf{Reference} & \textbf{\makecell{Spatial \\ Granularity}} & \textbf{\makecell{Temporal \\ Granularity}} & \textbf{\makecell{Forecasting \\ Types}} & \textbf{\makecell{Predictive \\ Model}} & \textbf{\makecell{Contextual \\ Inputs}} \\
\hline
\cite{15_lin2018predicting} & station& hour & deterministic & GCN with Data-driven Graph Filter & Temporal \\
\cite{liang2023cross} & station& 4-hour & deterministic & Multi-Relational GNN & Temporal, Public Transport \\
\cite{xu2018station} & TAZ & 10/15/20/30min & deterministic & LSTM & Temporal, Weather, Land Use \\
\cite{feng2024adaptive} & station& 20/40/60min & deterministic & \makecell{GCN with Group Multi-head Self-Attention} & Temporal, Weather, Land Use \\
\cite{22_xu2023multi} & station& hour & probabilistic & Transformer-Encoder-based Neural Process & Temporal, Censored Demand \\
\cite{13_gammelli2022predictive} & station & 15/30/40min & probabilistic & Variational Poisson RNN & Temporal, Weather, Censored Demand \\
\textbf{This study} & \textbf{station} & \textbf{15min} & \textbf{probabilistic} & \textbf{Transformer-based T-STAR} & \textbf{\makecell{Temporal, Weather, Facility, \\ Public Transport, Censored Demand}} \\
\hline
\end{tabular}%
}
\label{tab:forecasting_overview}
\end{table}


Table \ref{tab:forecasting_overview} summarizes key studies closely aligned with this work on short-term bike-sharing demand forecasting, categorizing them by spatial and temporal granularity, forecasting type, the adopted predictive model, and the contextual inputs considered. While the field has seen significant progress, especially in moving from coarse, deterministic models to finer-grained and probabilistic approaches, several important research gaps still remain unaddressed:

\begin{enumerate}
    \item \textit{Lack of structured inclusion of contextual factors}: While weather conditions and temporal characteristic of demand patterns are often considered in demand forecasting for bike-sharing services, existing models often do not systematically incorporate dynamic contextual information, such as the station facility properties (i.e., its holding capacity and the multimodal connectivity to other mobility services \citep{liang2023cross}) and the disruptions arising from censored demand \citep{22_xu2023multi, 13_gammelli2022predictive} or external latent events which are not directly observable in available data.

    \item \textit{Challenges at high spatio-temporal resolution}: Existing studies vary widely in spatial and temporal resolution. While some models address station-level forecasting \citep{liang2023cross, 22_xu2023multi}, few are equipped to handle ultra-short-term (e.g., 15-minute) predictions at such fine spatial resolution, which are more prone to noise, sparsity, and intermittency \citep{13_gammelli2022predictive}. It highlights the need for forecasting methods capable of generating accurate and robust predictions under high-resolution, volatile conditions. This is essential for supporting effective and proactive real-time operations in bike-sharing systems.

    \item \textit{Limited adoption of probabilistic forecasting methods}: Meanwhile, while probabilistic forecasts are essential for supporting risk-aware operations \citep{gast2015probabilistic, zhang2025stochastic}, most existing studies have focused on point predictions \citep{liang2023cross, feng2024adaptive}, with limited exploration of uncertainty quantification \citep{22_xu2023multi,13_gammelli2022predictive}. 
    As a result, current models often fall short in answering operationally relevant questions under uncertain conditions, such as the likelihood of bike availability or expected waiting time. 
    \textcolor{black}{While probabilistic forecasting has advanced across the broader urban mobility landscape, dock-based systems remain distinct. Unlike the spatial fluidity of ride-hailing or the schedule-dependency of transit, dock-based micro-mobility is fundamentally constrained by fixed station capacities. This physical constraint induces censored demand, where true user intent is obscured by station stockouts or saturation. Consequently, there is a critical need for predictive methods designed to reconcile these localized constraints with the sparse, zero-inflated patterns inherent in high-granularity station data for shared micro-mobility. Furthermore, current research lacks efficient frameworks capable of maintaining accuracy amidst high volatility while simultaneously capturing inter-modal dependencies. Bridging this gap requires models that ensure long-term generalizability across evolving service landscapes and enable real-time adaptation for dynamic uncertainty reassessment \citep{cheng2024recent}.}


\end{enumerate}

To address these gaps, we introduce transformer-based T-STAR, a two-stage framework for probabilistic, station-level demand forecasting in dock-based bike-sharing systems. 
This forecasting framework operates at a 15-minute temporal resolution, integrating a rich set of contextual features to capture the complex demand dynamics under real-time conditions. 

\section{Problem Formulation}
\label{chapter:problem_formulation}

This section presents the probabilistic demand forecasting problem for dock-based bike-sharing services. We begin by outlining the problem context, followed by key definitions of this study. Table \ref{tab:notations} provides an overview of the notations used in this study.

\begin{table}[h!]
\centering
\caption{A list of the key notations used throughout this study.}
\label{tab:notations}
\begin{tabular}{c|p{13cm}}
\hline
\textbf{Notation} &  \textbf{Description} \\ \hline
$i \in I$ & $i$ denotes a dock-based bike-sharing station; $I$ denotes the stations within the service network. \\ \hline
$\eta$    & A fixed time interval used to discretize the timeline of the forecasting problem. It determines the temporal resolution at which demand is aggregated and predictions are made (e.g., $\eta = 15$ minutes or $\eta = 1$ hour).\\ \hline
 $t$ & The general index for time intervals, denoting the temporal order on a demand time series. There are in total $T$ intervals in the time series generated from data.\\ \hline
  $y^i_t$ & The number of pickup or drop-off demand observed at station $i$ during the $t$-th time interval. \\ 
          & The target demand time series for station $i$ is $y^i = (y^i_1, y^i_2, \cdots, y^i_T), y^i_t \in \mathbb{Z}$. \\ \hline
  $X^i_t$  &  The contextual vector associated with station $i$ and time $t$. This includes:   static station-level attributes $x^i$, time-varying global  features $x_t$ shared across all stations, and dynamic, station-specific features $x^i_t$ that vary over time. \\ \hline
  $h$      & Hour index on the hourly aggregated demand time series. \\ \hline
  $q$      & Quarter index on the per-15-minute aggregated demand time series. \\ \hline
  $H$ & The forecasting horizon of a model, i.e. the number of future time intervals to be predicted. \\ \hline
  $V$ & The historical look-back window of the forecasting model, which defines the number of past time steps considered as input for the predictions.\\ 
  \hline 
\end{tabular}
\end{table}

\subsection{Dock-Based Bike-Sharing Services}

A dock-based bike-sharing service operates through a network of stations $I$, each is identified by a unique ID. Each station $i \in I$ is equipped with a fixed number of parking docks $C^i$, defining its maximum bike-holding capacity. At any time $t$, the operational status of a station is characterized by two dynamic metrics, fleet availability and parking availability.
Fleet availability represents the number of bikes available for pick-up, bounded between 0 and the station's hold capacity $C^i$. Parking availability indicates the number of vacant docks at the station, also ranging from 0 to $C^i$. 

Dock-based bike-sharing services typically provide 24/7 on-demand access for users to pick up and drop off bikes. Users may approach a station from various origins, such as shopping centers, campuses, or by transferring from other transit services as part of a multimodal journey. A typical trip starts when a user unlocks a bike at one station and ends when the bike is returned at another. Historical records capture completed trips, including timestamps and the origin and destination stations.
A pickup is not possible when no bikes are available, and a drop-off can be denied when the parking is full. In such situations, users may reroute to an alternative station or cancel their trip. These unfulfilled trips are not logged in the historical data, which only records completed journeys, limiting insights into unmet demand.

As reviewed in section \ref{subsect:contextual_features}, bike-sharing demand is influenced by a combination of internal and external factors. These include recurrent usage patterns, weather conditions, and the operations of nearby public transit services. In this study, contextual features are integrated into the forecasting framework and categorized based on whether they are exogenous to the system and whether they are dynamic or static.
Specifically, to account for multimodal connectivity, a bike-sharing station $i$ is considered connected to a nearby public transit service $m$ if its location $L^i$ is within a threshold distance $\nu$ \citep{ma2018understanding}. 
\subsection{Probabilistic Short-Term Demand Forecasting for Bike-Sharing Services}
       
To ensure efficient service and customer satisfaction, bike-sharing providers require timely forecasts of both pickup and drop-off demand at each station. These activities directly impact station-level inflow and outflow, dynamically affecting fleet availability in the short term.
Robust operational planning further benefits from probabilistic forecasts that quantify uncertainty in future demand. In this study, we propose a probabilistic forecasting approach that integrates historical trip records with station-level attributes and exogenous contextual features to predict short-term pickup and drop-off demand at individual stations.

We define a fixed time interval $\eta$ (e.g., 15 minutes or 1 hour) as the temporal resolution of the forecasting task. This interval determines both the aggregation of demand data and the frequency of prediction updates.
For each station $i$, the historical trip records are processed to generate a demand time series $y^i_t = (y^i_1, y^i_2, \cdots, y^i_T)$, where each $y^i_t \in \mathbb{Z}$ represents the number of demand (either pickups or drop-offs) that occurred during the $t$-th time interval. Depending on the specific forecasting objective, $y^i_t$ may refer to the sequence of pickup counts or drop-off counts.
Each target value  $y^i_t$ is associated with a multivariate feature vector $X^i_t$, representing contextual information available at time $t$ to support forecasting.


The goal of the probabilistic (multi-step ahead) demand forecasting problem is to estimate the conditional distribution of future demand over the prediction horizon, given historical observations and available contextual information. 
Given a future horizon $H$, the task is to estimate,
\begin{equation}
    p_{\mathcal{D}}(y_{t+1:t+H} | X_{t-V+1:t}, y_{t-V+1:t}, X^*_{t+1:t+H}),
\end{equation}
where,
\begin{itemize}
    \item $y_{t+1:t+H}$ denotes the demand to be predicted over the next $H$ time intervals;
    \item $X_{t-V+1:t}$ and $y_{t-V:t}$ represent the historical demand and contextual features observed during the past $V$ intervals (i.e., the observable input window);
    \item $X^*_{t+1:t+H}$ denote the knowledge about the contextual information over the future horizon. In this study, we assume temporal information (like day-of-the-week and hour-of-the-day) and the estimated weather conditions are available for the future forecasting horizon as inputs.
\end{itemize}
Probabilistic forecasting models assume that future demand follows an underlying distribution $\mathcal{D}$ and aim to characterize this distribution rather than predict a single point estimate. By providing a distributional view of future outcomes, such as predicted quantiles or prediction intervals, probabilistic forecasting captures the inherent uncertainty in demand. These distributional insights enable uncertainty-aware decision-making, which is critical for robust planning and can be directly integrated into stochastic optimization frameworks.




\section{Methodology}
\label{chapter:methods}

This section presents the Transformer-based Two-stage Spatial and Temporal Adaptive contextual Representation (T-STAR) developed for short-term demand probabilistic forecasting in dock-based shared micro-mobility systems. 

T-STAR builds on the hierarchical modeling of global and local random effects in multilevel time series forecasting \citep{wang2019deep}, as well as recent advances in hierarchical prediction frameworks \citep{chen2020detecting, khalesian2024improving}. 
Its two-stage structure is designed to disentangle stable demand patterns from short-term fluctuations. 
At each stage of T-STAR, a multivariate time series Transformer models sequential demand in relation to the selected contextual features.
Through this hierarchical integration of contextual information, T-STAR effectively captures both global structures and local dynamics, enabling accurate and uncertainty-aware forecasts.

Unlike multi-layer forecasting models such as the Temporal Fusion Transformer or ResNet-based architectures, which operate on time series data at a single temporal resolution \citep{lim2021temporal, shen2020novel}, T-STAR introduces granularity-aware hierarchical supervision, where forecasts at each temporal level are trained independently before being fed to the downstream. This design reduces the risk of error propagation from noisy high-resolution time series and allows fine-grained predictions to benefit from the stability of aggregated representations learned at coarser scales, enhancing performance under sparse and volatile demand conditions.

Section \ref{sect:two-stage} details the architecture and design rationale of the two-stage framework, explaining how contextual embeddings from Stage 1 are refined in Stage 2 to improve predictive accuracy. Section \ref{sect:transformer} introduces the Time Series Transformer, the selected base predictor in T-STAR, and explains how it generates probabilistic demand predictions while accommodating multivariate inputs. 



 \subsection{A Two-stage Spatial and Temporal Adaptive contextual Representation (T-STAR) Framework}
\label{sect:two-stage}

The T-STAR framework addresses the challenge of forecasting both stable demand patterns and short-term fluctuations by decomposing the task into two sequential stages. This two-stage design allows the model to align the temporal granularity of input features with their corresponding predictive targets, enabling the learning of more meaningful contextual representations while improving both predictive accuracy and computational efficiency.
Specifically, Stage 1 models the underlying regular demand trends by leveraging coarse-grained contextual features, such as hourly weather conditions, time-of-day, and day-of-the-week features, to generate probabilistic estimate of hourly demand. Stage 2 then models short-term variations by focusing on localized deviations from this baseline, incorporating fine-grained temporal inputs and dynamic signals of recent supply-demand imbalances. This separation is designed to mitigate the impact of data sparsity in high-frequency time series and support more targeted use of context at each stage.

\begin{figure}[h!]
    \centering
    \includegraphics[width=0.8\linewidth]{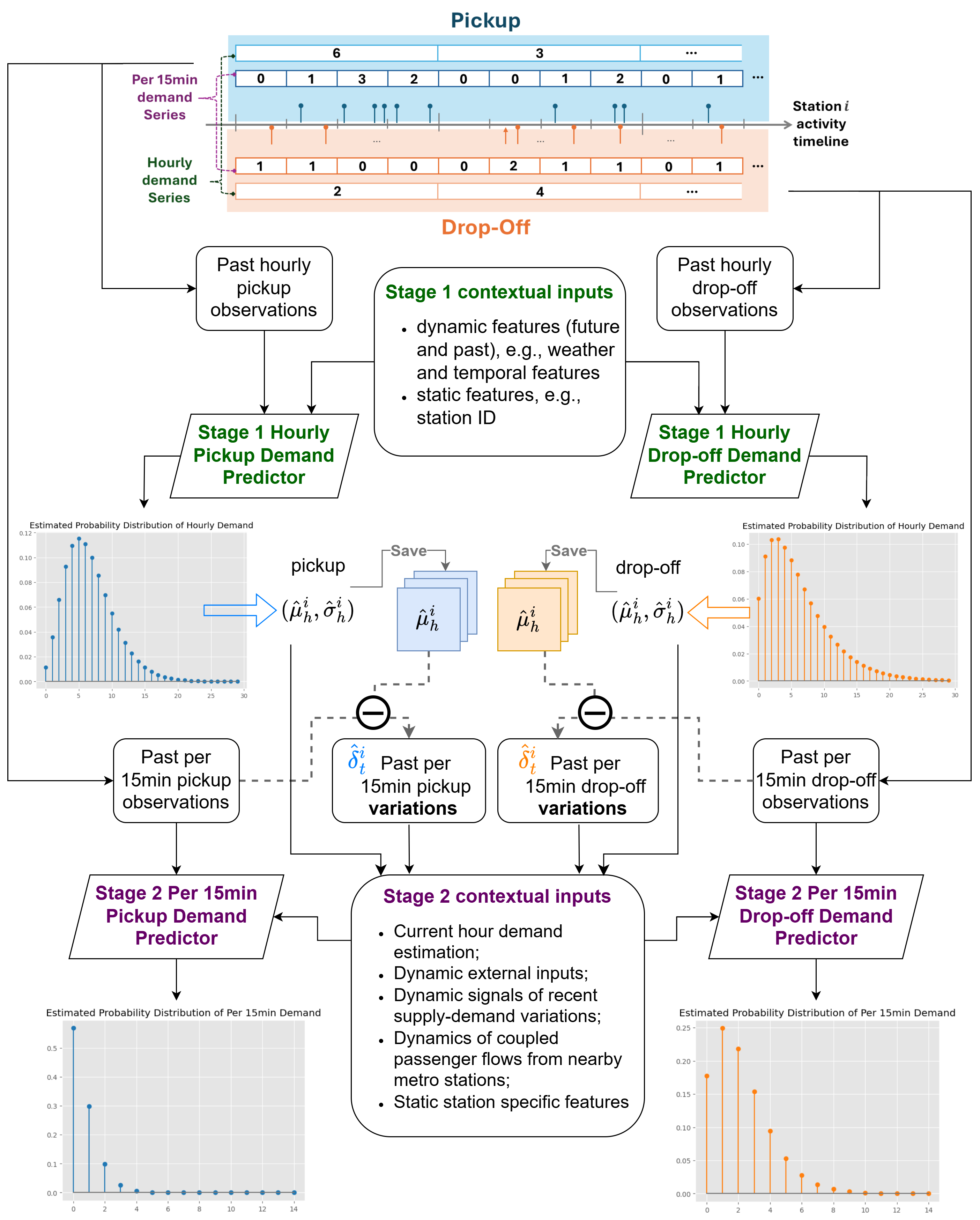}
    \caption{Pickup and drop-off demand forecasting pipeline of Two-stage Spatial and Temporal Adaptive contextual Representation (T-STAR).}
    \label{fig:T_STAR_structure}
\end{figure}

Figure \ref{fig:T_STAR_structure} illustrates the hierarchical feature processing pipeline of the T-STAR framework, which generates probabilistic pickup and drop-off demand forecasts for the next 15-minute interval based on historical observations and multivariate contextual inputs for each station. 

\subsubsection*{Stage 1: Modeling Hour Demand Patterns}

In the first stage of the T-STAR framework, the model estimates the underlying regular demand trend by forecasting future hourly demand based on recent demand history and coarse-grained contextual information. Specifically, the model takes as input the past $V_1$ observations on the hourly demand sequence $y^i_{h-V_1+1:h}$ and the associated contextual features $X^i_{h-V_1+1:h}$ for station $i$, along with known or externally estimated future features $X^i_{h+1:h+H_1}$ over the forecasting horizon of length $H_1$. The subscript in $V_1$ and $H_1$ indicates that these hyperparameters pertain to Stage 1 of the T-STAR framework.
The output is an estimated conditional distribution over the future hourly demand for station $i$, denoted by $\hat{y}_{h+1:h+H_1}^{i}$, and is defined as:
\begin{equation}
\hat{y}^{i}_{h+1:h+H_1} \sim \mathcal{F}_{1}\left( y_{h-V_1+1:h}^{i},\ X_{h-V_1+1:h}^{i},\ X^{*i}_{h+1:h+H_1} \right),
\end{equation}
where $\mathcal{F}_{1}$ is the Stage 1 forecasting model. Note that, T-STAR framework is model-agnostic and allows for different base predictors, such as Time Series Transformer and XGBoost.
$y_{h-V_1+1:h}^{i}$ is the historical hourly demand series. $X_{h-V_1+1:h}^{i}$ includes temporal and environmental covariates aligned with those observations. And $X^{*i}_{h+1:h+H_1}$ contains forward-looking contextual features used to inform the forecast, which are either known in advance or estimated from external sources.

For the experiments in this study, we set $V_1 = 24$ to incorporate one full day of past hourly observations and $H_1 = 1$ to predict demand for the next hour. As the targeted forecasting problem in this study focuses on predictions for the next 15-minute, predicting the one-hour ahead demand from Stage 1 is sufficient. The value of $H_1$ can be adjusted as needed for longer-term forecasting problems.
To capture regular demand patterns, the contextual factors, denoted by $X_{h-V_1+1:h}^{i}$ and $X^{*i}_{h+1:h+H_1}$, include temporal variables hour-of-the-day, day-of-the-week, and a binary indicator for national holidays. These features help describe the cyclical demand trends observed in historical data. In addition, hourly weather information (both past observations and one-hour-ahead forecasts) covering wind speed, precipitation, and temperature, are included to account for environmental influences on mobility behavior.


\subsubsection*{Intermediate Demand Variational Signals Generation}

To generate dynamic demand variation signals, we compare fine-grained (15-minute) demand observations with coarse-grained (hourly) demand expectations derived from Stage 1 predictions. This process unfolds in three steps, as explained in the following.

Firstly, for each station $i$, the predicted mean $\hat{\mu}^i_{h+1}$ and standard deviation $\hat{\sigma}^i_{h+1}$ are extracted from the Stage 1 forecast of hourly demand for the next hour. 
The mean $\hat{\mu}^i_{h+1}$ represents the expected demand under the current temporal and contextual conditions (e.g., time of day and weather).
As hourly forecasts are generated over time, these values are stored as sequences, forming a dynamic baseline of regular demand expectations and supporting the identification of short-term demand deviations.

Secondly, as hourly forecasts are continuously updated, we retrospectively compare them with the actual per 15-minute demand observations that occurred during the same hour. Say the current 15-minute interval is $t$, for a previous 15-minute interval indexed by $t-q'$, we identify its corresponding hour $h'$  and compute the estimated 15-minute demand from Stage 1 as:
\begin{equation}
    \tilde{\mu}^i_{t-q'} = \frac{1}{4} \hat{\mu}^i_{h'}, 
\end{equation}
assuming demand is uniformly distributed across the four quarters of hour $h'$.

Thirdly, the dynamic demand variation signal is then calculated as the difference between the actual observation and the scaled hourly estimate:
\begin{equation}
\delta^i_{t-q'} = y^i_{t-q'} - \tilde{\mu}^i_{t-q'} = y^i_{t-q'} - \frac{1}{4} \hat{\mu}^i_{h'}.
\end{equation}
By repeating this process over previous time steps, we retain a sequence of demand variation signals $ \{ \delta^i_{0}, \cdots, \delta^i_{t-1}\}$, which captures recent deviations from the expected demand. This sequence is a key dynamic input in Stage 2.

\begin{figure}[h!]
    \centering
    \includegraphics[width=0.8\linewidth]{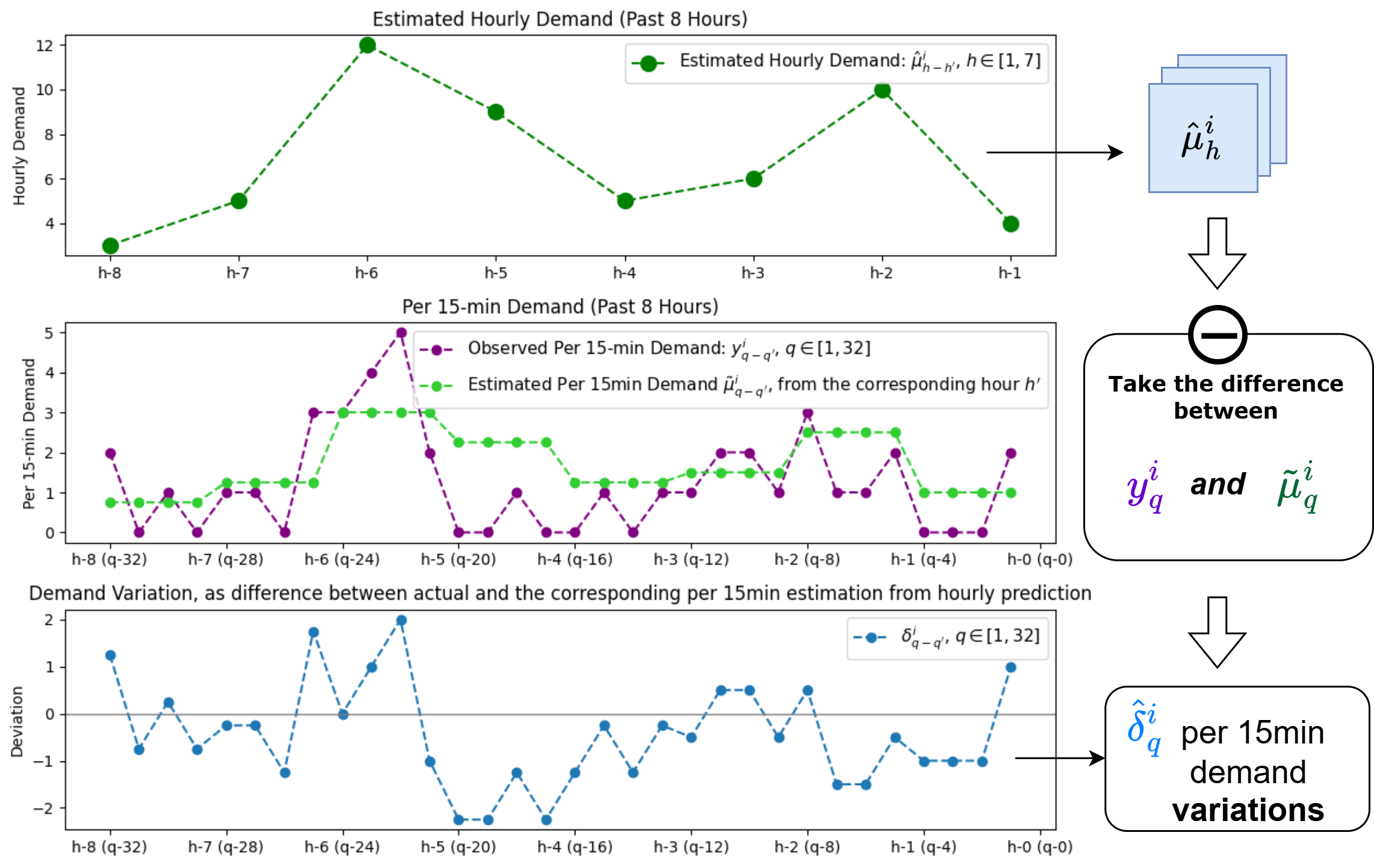}
    \caption{Illustration of the generation process of intermediate demand variational signals.}
    \label{fig:delta_diagram}
\end{figure}

Figure \ref{fig:delta_diagram} illustrates the process of generating dynamic demand variation signals. The first panel from the top shows the predicted hourly demand $\hat{\mu}^i_{h}$ over the past 8 hours. The middle panel visualizes the actual per 15-minute demand values $y^i_{q}$ alongside their corresponding hourly-derived estimates $\tilde{\mu}^i_{q}$, for each short-term interval index by $q$. The bottom panel depicts the resulting demand variation signals $\delta^i_{q}$, capturing localized fluctuations relative to regular hourly expectations.

\subsubsection*{Stage 2: Capturing Short-Term Demand Variations}

The second stage of the T-STAR framework focuses on capturing high-frequency fluctuations in short-term demand that deviate from the regular hourly patterns estimated in Stage 1. This refinement step enables adaptive 15-minute forecasts by incorporating additional signals not present in the coarse-grained predictions. 
Stage 2 models future short-term demand by integrating three primary sources of influence: (1) station-specific short-term demand patterns; (2) localized demand fluctuations driven by latent or unobserved disruptions, captured through recent demand variation signals; and (3) contextual factors arising from the surrounding mobility environment. 

To establish a baseline understanding of current demand, the Stage 2 model first leverages recent station-level demand observations and temporal features. Specifically, the previous $V_2$ per 15-minute demand values $y^i_{t-V2+1:t}$ are used to capture short-term autoregressive patterns that reflect local demand trends at station $i$.
In addition, the predicted hourly demand distribution from Stage 1 provides a coarse-grained expectation of demand. The predicted mean $\hat{\mu}^i_{t}$ offers a reference level for typical demand under prevailing contextual conditions considered in Stage 1, while the standard deviation $\hat{\sigma}^i_{t}$ quantifies the associated uncertainty. 
These values are passed into Stage 2 as compact summaries of the expected demand, without the need to reintroduce the contextual inputs already processed in Stage 1. Finally, demand seasonality indicators, hour-of-the-day and day-of-the-week, are included to identify the seasonal short-term demand patterns. 

In addition to recent demand observations, the Stage 2 model incorporates demand variation signals to account for sudden shocks or irregular changes in short-term demand. The signals, denoted by $\delta^i_{t-V_2+1:t}$, capture temporal clusters of deviations caused by latent events, such as concerts, local disruptions, or anomalies in the broader mobility system. They are often not directly observable by the bike-sharing systems but may significantly alter user behavior. Moreover, demand variation signals act as indirect indicators of operational constraints, including censored demand resulting from zero fleet availability or limited parking space at a station. For example, a surge in pickups unaccompanied by corresponding drop-offs may indicate a depletion of available vehicles, potentially suppressing subsequent pickup activity despite continued user demand. Since real-time fleet availability data is often unavailable, the model leverages recent variation patterns to infer such imbalances. To support this inference, the Stage 2 predictor incorporates the station's dock capacity $C^i$, along with both pickup and drop-off variation sequences $\delta^{i,p}_{t-V_2+1:t}$ and $\delta^{i,d}_{t-V_2+1:t}$, where superscripts $p$ and $d$ refer to pickup and drop-off demand.

In addition to temporal patterns and variation signals, the Stage 2 model incorporates contextual features that capture the station’s surrounding mobility environment. A key component is the interaction with nearby public transit, as shared micro-mobility services often serve as first- and last-mile connectors. To capture this multimodal dependency, the model includes real-time deviations in metro activity. For each 15-minute interval, we compute the difference between the observed number of check-ins or check-outs at the nearby metro station (within predefined distance threshold $\nu$) and their historical average for that interval, conditioned on the same hour-of-the-day and day-of-the-week. These deviations serve as proxies for irregular shifts in multimodal travel demand and help the model adjust to sudden changes of usage from the connecting public transport services. Specifically, check-out deviations $\delta_t^{i,out}$ are used for pickup demand forecasting, while check-in deviations $\delta_t^{i,in}$ inform drop-off demand forecasting.

In summary, the Stage 2 predictor estimates the conditional distribution of the next 15-minute demand at station $i$ and time $t$, incorporating recent demand trends, the predicted demand expectations from Stage 1, recent demand variation signals, and contextual features. Formally, the predicted demand distribution is given by:

\begin{equation}
\hat{y}^i_{t+1:t+H_2} \sim \mathcal{F}_{2} \left(
\underbrace{y^i_{t-V_2+1:t}}_{\text{recent demand}},\ 
\underbrace{\delta^{i,p}_{t-V_2+1:t},\ \delta^{i,d}_{t-V_2+1:t}}_{\text{past variation signals}},\ 
\underbrace{\hat{\mu}^i_{t+1:t+H_2},\ \hat{\sigma}^i_{t+1:t+H_2}}_{\text{Stage 1 estimations}},\ 
\underbrace{X^i_{t-V_2+1:t},\ X^{*i}_{t+1:t+H_2}}_{\text{contextual features}}
\right),
\end{equation}
where $\mathcal{F}_{2}$ is a predictor designed to process multivariate input streams and produce a probabilistic forecast of short-term demand.

In this study, which focuses on forecasting dock-based bike-sharing demand for the next 15-minute interval, the look-back window $V_2$ of the Stage 2 predictor is set to 24. This allows the model to access demand and contextual information from the past 6 hours (i.e., 24 intervals of 15 minutes). The prediction horizon $H_2$ is set to 1, as the task focuses on one-step-ahead forecasting of demand for the next 15-minute interval. As with the Stage 1 predictor, the values of $V_2$ and $H_2$ can be adjusted based on data availability and specific operational requirements. 

Regarding external contextual inputs, the feature sets $X^i_{t-V_2+1:t}$ and $X^{*i}_{t+1:t+H_2}$ include the per 15-minute metro check-in or check-out flow variation signals ($\delta_t^{i,out}$ or $\delta_t^{i,in}$) from metro stations located near the target bike-sharing station. If a bike-sharing station is connected to multiple metro stations, the variation signal aggregates the passenger flow deviations from all nearby stations for the same interval. In addition to these multimodal signals, the contextual inputs also include temporal seasonality features such as hour-of-the-day and day-of-the-week, as well as the station’s docking capacity $C^i$.

\subsection{Multivariate Time Series Transformer for Probabilistic Demand Forecasting}
\label{sect:transformer}

\subsubsection*{Time Series Transformer (TST)}

Originally introduced for natural language processing \citep{vaswani2017attention}, Transformer architectures have recently shown strong performance in time series forecasting tasks, thanks to their ability to capture long-range temporal dependencies and process multivariate inputs efficiently \citep{zerveas2021transformer}. For the proposed T-STAR framework, TST is selected as the base predictor for both Stage 1 and Stage 2 models. TST is trained as a global model across all stations, learning station-specific representations while leveraging the shared patterns of demand, which makes it particularly suitable for large-scale bike-sharing systems.

\begin{figure}[h!]
    \centering
    \includegraphics[width= 0.65 \linewidth]{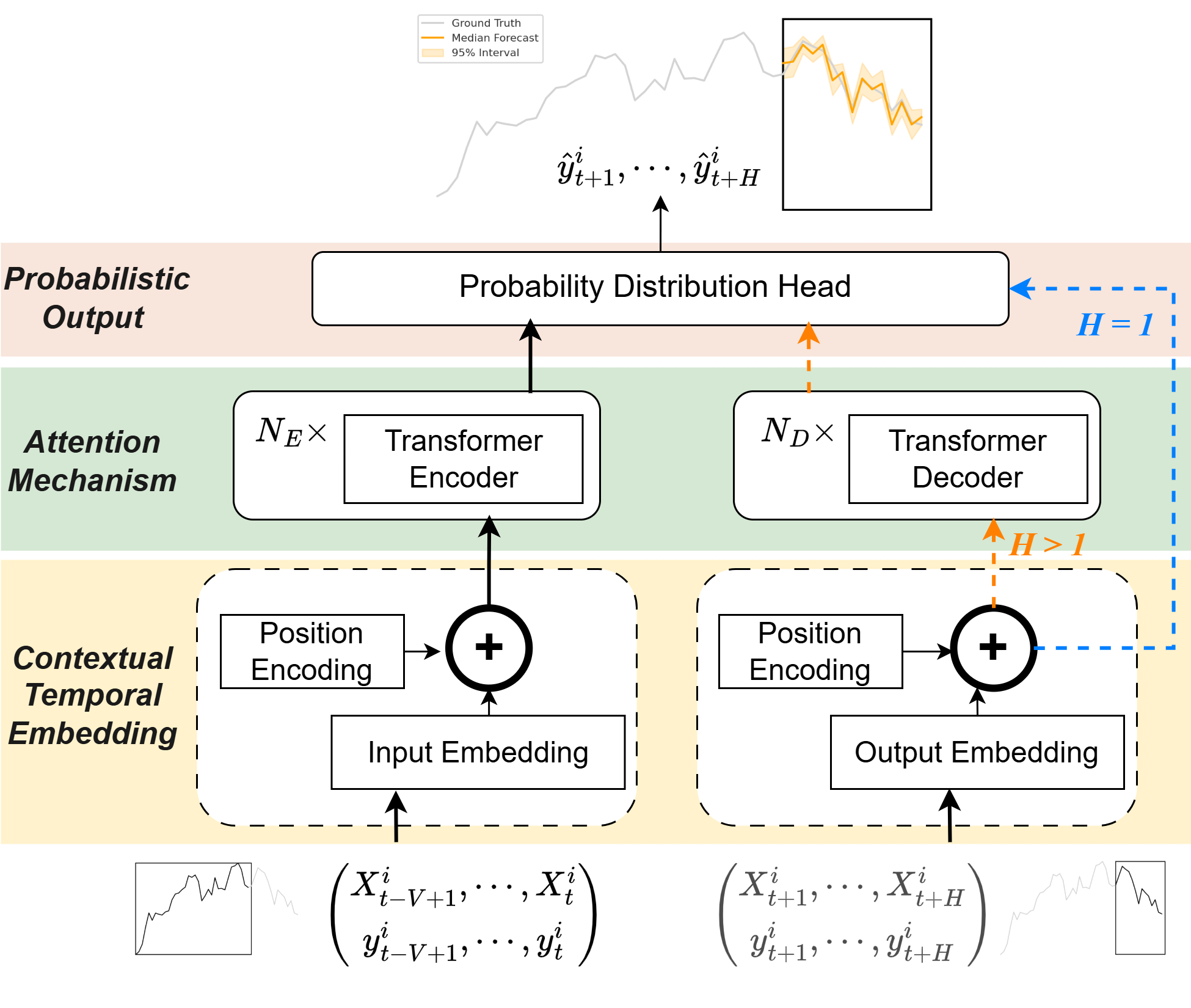}
    \caption{Architecture of a Transformer-based probabilistic forecasting model. \textcolor{black}{Note: For one-step forecasting ($H=1$), the decoder branch is inactive, and the encoder output is passed directly to the distribution head.}}
    \label{fig:transformer_architecture}
\end{figure}

Illustrated in Figure \ref{fig:transformer_architecture}, the architecture of TST consists of a contextual embedding layer, self-attention–based encoder and decoder blocks, and a probabilistic output head. The model takes as input a context window of past demand values and associated multivariate features and predicts a probabilistic forecast for the next time step(s). The input and output sequences are first transformed into fixed-dimensional embeddings using learnable projection layers and positional encodings. These embeddings are then processed by the encoder or decoder modules, which are composed of $N_E$ and $N_D$ stacked encoder and decoder blocks, respectively. Each block contains a multi-head self-attention mechanism followed by position-wise feed-forward sublayers \citep{vaswani2017attention}. While the encoder summarizes historical patterns and contextual influences, The decoder enables multi-step forecasting ($H>1$) by autoregressively attending to prior predictions at each step \citep{vaswani2017attention}. 
Unlike RNNs, which process sequences sequentially due to their recursive structure, Transformers leverage self-attention mechanisms to process the entire input sequence in parallel. This eliminates sequential computation during training, enabling efficient mini-batching and accelerating training on large datasets.

\textcolor{black}{While the T-STAR framework incorporates a decoder for multi-horizon flexibility, it is important to note that for all experiments in this study where $H=1$, the decoder is bypassed. This reduces the computational overhead during training and inference, effectively leveraging an encoder-only configuration to capture the temporal dependencies on the time series.}

The Transformer architecture offers several key advantages for demand forecasting, including the ability to model complex temporal dependencies without recurrence, enabling parallel training and scalability for large-scale systems \citep{vaswani2017attention, zerveas2021transformer}. Its global learning framework enables both zero-shot generalization, where forecasts can be produced for new stations or service areas without additional training, as well as few-shot adaptation through lightweight fine-tuning over the pretrained Transformer. This flexibility is particularly valuable for fast-growing shared micro-mobility services, where new locations are frequently introduced. Moreover, the model can be incrementally updated using recent data, improving adaptability to evolving usage patterns and external conditions of the bike-sharing systems. When equipped with a distribution head (e.g., Negative Binomial), it enables uncertainty-aware probabilistic forecasting.


\subsubsection*{Contextual Temporal Embeddings}

In addition to modeling temporal dependencies through past demand observations $y^i_{t-V+1:t}$ the contextual embedding layer of TST also learns contextual representations from associated historical covariates $X^i_{t-V+1:t}$ and future covariates $X^{*i}_{t+1:t+H}$ with known or externally estimated values. Here, the subscript $t$ refers to the current time point at which the prediction is generated, while $\tau$ is used as a generic index that may refer to any point of time within the input window.
To support generalization and structured modeling, contextual inputs are categorized into three types: the static station-level attributes $x^i$, the global time-varying features shared across all stations $x_\tau$, and station-specific, time-dependent features $x^i_\tau$. Together, at any given time point $\tau$, the contextual input at time $\tau$ for station $i$ is denoted as $X^s_\tau = \{ x^i, x_\tau, x^i_\tau\}$. This structured input enables the model to represent both global and local influences on demand across stations.

Our forecasting approach uses a global Transformer model trained on data from all stations. To capture station-specific characteristics, the model uses static features $x^i$, which include both the station ID (as a categorical feature) and operational attributes such as station capacity $C^i$ from the Stage 2 model. 
The station ID is embedded through a learnable embedding layer, enabling the model to infer unique historical demand patterns associated with each station. These static inputs collectively produce a latent station representation $h^i \in \mathbb{R}^i$, reflecting the historical behavior and demand patterns of each station. If two stations exhibit similar demand dynamics and responses to contextual factors, their learned embeddings will also be similar, allowing the model to generalize effectively across the network while preserving station-specific nuances.

Dynamic global features $x_\tau$, including weather conditions, hour-of-the-day and day-of-the-week, are encoded through a dense layer into a shared temporal embedding $h_\tau \in \mathbb{R}^s$. These inputs capture periodic and seasonal variations common across the network. Conversely, local dynamic features $x^i_\tau \in \mathbb{R}^p$, such as passenger flow variations and recent demand deviation signals, reflect the dynamic conditions specific to station $i$. 
\textcolor{black}{To enable the Transformer to simultaneously reason across different data scales, we fuse heterogeneous inputs into a unified feature space. Specifically, we concatenate the static station latent representation $h^i$, the global temporal embedding $h_\tau$, and station-specific dynamic features $x_\tau^i$ with recent demand observations $y^i_\tau$. This fusion, formalized in Equation \ref{eq:6}, ensures that the model's view of a single time step is informed by both long-term station characteristics and dynamic contextual conditions:}
\begin{equation}
\label{eq:6}
    h^i_\tau = [h^i, h_\tau, x^i_\tau, y^i_\tau]^T \in \mathbb{R}^{d}, 
\end{equation}
where $d = i+s+p+1$, with $i$, $s$, and $p$ representing the dimensions of the station embedding, global temporal embedding, and station-specific dynamic context, respectively.
This concatenated vector is then passed through a shared projection layer:
\begin{equation}
    z^i_\tau = h^i_\tau W_e + b_e,
\end{equation}
where $W_e \in \mathbb{R}^{d \times e}$ and $b_e \in \mathbb{R}^{e}$ to map the input into a fixed-size embedding vector that can be uniformly processed by the Transformer encoder.

In addition to contextual embeddings, the model also incorporates positional information to preserve the temporal order of observations. 
Unlike recurrent models, Transformers process all input tokens in parallel and lack an inherent notion of sequence order. 
To address this, positional encodings $PE_\tau \in \mathbb{R}^e$ are incorporated to inject position-specific information into the input representations \citep{vaswani2017attention}. These encodings are computed using the sinusoidal encoding scheme from \cite{vaswani2017attention}, allowing the Transformer to infer token order through deterministic position-dependent signals. Adding the positional embedding element-wisely to contextual embeddings $z^i_\tau $, the resulting embedding is,
\begin{equation}
    e^i_\tau = z^i_\tau + PE_\tau.
\end{equation}

\textcolor{black}{Transformer requires a structured temporal sequence to perform self-attention. Equation \ref{eq:9} formalizes the aggregation of these individual embeddings ($e^i_\tau$) into a sequence matrix $E_t^i \in \mathbb{R}^{V \times e}$ over the look-back window $V$. This structured temporal sequence serves as the primary input for the Transformer Encoder, providing the necessary global context for the model to capture long-range dependencies within the bike-sharing network:}
\begin{equation}
\label{eq:9}
    E^i_t = [e^i_{t-V+1}, \cdots, e^i_\tau, \cdots e^i_{t}]^T \in \mathbb{R}^{V \times e}.
\end{equation}

\subsubsection*{Probabilistic Output}

To support uncertainty-aware forecasting, the TST in T-STAR is trained to produce a predictive distribution rather than a single point estimate. Specifically, the model outputs the parameters of a Negative Binomial (NB) distribution at each prediction step. The NB distribution, also known as the Gamma-Poisson distribution, generalizes the Poisson model by introducing a dispersion parameter that allows the variance to exceed the mean. This property makes it particularly suitable for modeling short-term bike-sharing demand, which is discrete, non-negative, and low in volume but highly variable over 15-minute intervals.
The NB distribution has been widely adopted in probabilistic forecasting of discrete, intermittent demand. Prior studies highlight its suitability for highly variable customer demand \citep{salinas2020deepar, toubeau2018deep}. A detailed formulation of the NB distribution and its likelihood function is provided in \ref{appendix:NB}.

The Transformer's probabilistic output layer estimates the parameters of the Negative Binomial (NB) distribution, which are the mean and dispersion, for each future time step. To generate predictive uncertainty estimates, the model samples $N$ values from the predicted distribution, forming an empirical forecast distribution from which statistics such as the median and standard deviation can be derived. 
Model training is performed by minimizing the Negative Log-Likelihood (NLL) of the observed demand under the predicted NB distribution. This learning objective encourages the model to correctly predict the expected value of demand and to capture the uncertainty of demand. In a general form, the NLL loss is defined as:
\begin{equation}
    \mathcal{L}_{\text{NLL}} = -\log p_{\mathcal{D}}(y_t \mid \hat{\theta}_{t})
\end{equation}
where $y_{t}$ is the observed demand at time step $t$, $\mathcal{D}$ denotes the assumed predictive distribution (NB distribution in this study), and $\hat{\theta}_{t}$ represents the predicted distribution parameters.

\section{Case Study and Experimental Setup}
\label{chapter:case}




In the following, we introduce the case study and describe how the proposed contextual two-stage demand forecasting framework is calibrated to this setting. We then present the performance evaluation scheme, which assesses both point and probabilistic prediction accuracy. To provide a comprehensive view of model performance, we evaluate not only overall accuracy but also robustness across stations and time. Finally, we outline the design of benchmarking experiments, including a summary of the selected baseline models used for comparison.

\subsection{Case Study Description}
\label{subsect:case_study_and_model_calibration}
To evaluate the performance of the proposed short-term demand forecasting methodology in a real-world setting, we apply it to a case study of a dock-based bike-sharing system. The objective is to predict pickup and drop-off demand for each bike-sharing station in downtown Washington D.C. for the future 15-minute interval. To balance computational efficiency with high predictive accuracy, we adopt a global prediction model which simultaneously forecasts future pickup and drop-off demand across all stations in the network.

In our analysis, we utilize historical trip data from Washington D.C.’s Capital Bikeshare system, covering the period from October 1 to December 31, 2022\footnote{The dataset is publicly available via Capital Bikeshare's open data portal: \url{https://capitalbikeshare.com/system-data}}. On average, the system processes over 6,000 trips per day across 235 selected bike-sharing stations distributed throughout the city center of D.C.

\begin{figure}[h!]
    \centering
    \includegraphics[width=0.65\linewidth]{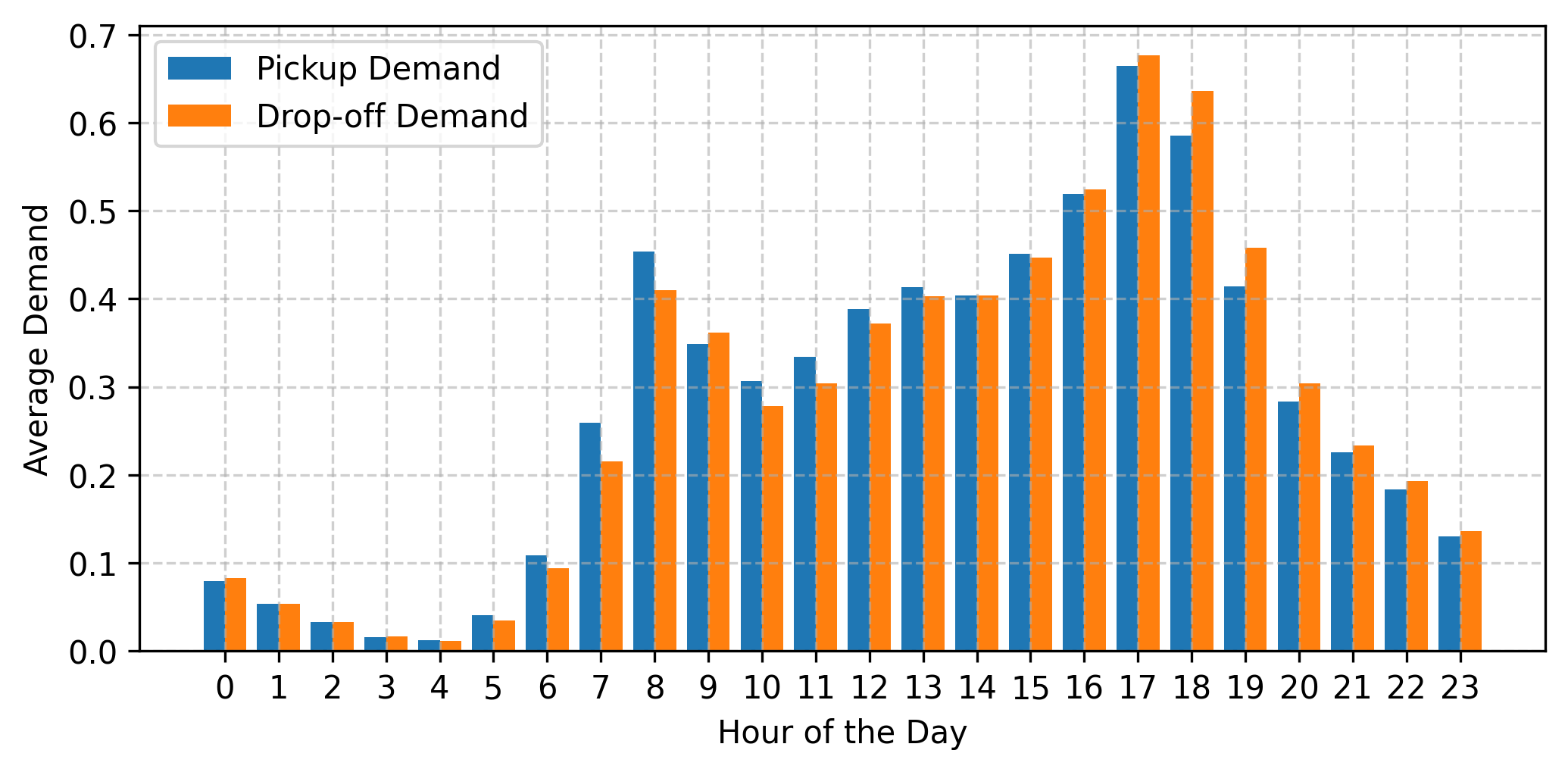}
    \caption{Hourly average bike pickup and drop-off demand (Washington D.C., Oct–Dec 2022).}
    \label{fig:hourly_avg_demand}
\end{figure}

The demand pattern of bike-sharing services is often characterized by strong seasonality \citep{eren2020review}. As shown in Figure \ref{fig:hourly_avg_demand}, the hourly demand exhibits significantly higher activity during daytime hours, from 8:00 a.m to 8:00 p.m. The pattern clearly reflects commuter behavior, with pronounced peaks during the morning (8:00–10:00 a.m.) and early evening (4:00–7:00 p.m.) rush hours. Moreover, on weekdays, these commuting peaks appear consistently, indicating regular usage by urban commuters. In contrast, weekend demand is more evenly distributed throughout the day, suggesting a shift toward leisure-oriented usage. Furthermore, public holidays, while reducing commuting-related trips, often stimulate increased demand at bike-sharing stations near recreational areas and tourist attractions \citep{palaio2021multicity}. A similar pattern is observed in the bike-sharing demand data from Washington D.C.

The demand time series for each BSS is constructed by aggregating the number of pickups and drop-offs into 15-minute intervals over the entire historical period. This results in 96 demand observations per day for each station. Temporal features such as hour-of-the-day and day-of-the-week are extracted from the timestamp of each interval to capture daily and weekly seasonal patterns. Additionally, a binary indicator is created to denote whether a given interval falls on a public holiday.
To account for weather influences, historical hourly temperature, precipitation, and wind speed data are incorporated to support the predictive analysis \citep{openmeteo}. To capture the interaction between public transit and bike-sharing demand, we incorporate metro passenger flow data, defined as the number of entries and exits at metro stations per 15-minute interval. This information is derived from boarding and alighting records provided by the Washington Metropolitan Area Transit Authority (WMATA) Metrorail system.
To link metro activity to bike-sharing demand, we define a proximity threshold of 300 meters, approximately a 5-minute walk, following the guideline in \cite{ma2018understanding}. A bike-sharing station is considered connected to a metro station if it falls within this distance. Based on this criterion, a total of 63 BSS are identified as being connected to at least one metro station. Finally, the docking capacity of each BSS is retrieved from Capital Bikeshare dataset and imported as a static feature.

\begin{figure}[h!]
    \centering
    \includegraphics[width=0.5\linewidth]{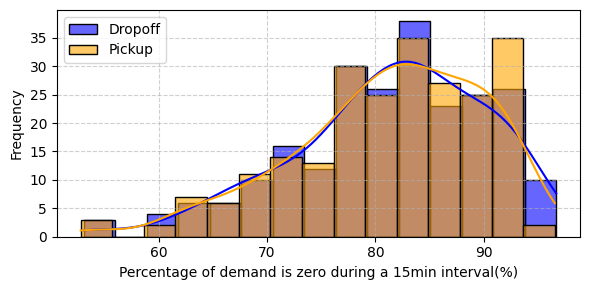}
    \caption{Distribution of zero-demand percentages in the historical 15-minute intervals across 235 bike-sharing stations in Washington D.C case study.}
    \label{fig:zero_demand}
\end{figure}

In the analysis of the demand time series at the 15-minute resolution, we observe that a large proportion of data points are zero, particularly during early morning and late evening hours.
The histogram in Figure \ref{fig:zero_demand} shows the distribution of zero-demand frequencies of 15-minute intervals across all stations. For most stations, zero demand occurs in over 70–90\% of the 15-minute intervals, for both pickup and drop-off requests. This highlights the high degree of zero-inflation in the station-level demand time series for this case study.

Demand data from the first 70 days (\textcolor{black}{October 2 to December 10}, 2022) are used for model training, while the remaining 20 days are reserved for testing and evaluating predictive performance. This setup reflects a practical trade-off given the limited dataset size. The short validation window minimizes data withheld from training while still enabling basic hyperparameter tuning. The validation day was selected to reflect typical weekday demand conditions.


The proposed T-STAR forecasting framework is applied to capture both multivariate long-term patterns and short-term fluctuations in future demand at a 15-minute resolution. It does so by integrating diverse factors from the external environment and the recent supply-demand dynamics of the bike-sharing system itself. These features either describe the long-term demand trends or drive localized short-term variations. 
 

 
In the first stage, demand is aggregated into hourly intervals. An hourly prediction model is then trained using these aggregated time series, along with lower-frequency contextual features such as weather conditions, public holidays, and seasonal indicators (e.g., hour-of-the-day and day-of-the-week). This stage should produce reliable hourly demand estimates, which serve as coarse-grained inputs for the second stage.

In the second stage, a short-term forecasting model predicts demand at 15-minute intervals within each hour. This model incorporates multiple sources of information: (1) recently observed 15-minute demand, (2) seasonal indicators (e.g., hour-of-the-day and day-of-the-week), (3) probabilistic hourly predictions from the first stage, and (4) high-frequency contextual features updated every 15 minutes.

In the Washington D.C. case study, the third source mentioned, i.e. the hourly predictions from the first-stage model, are represented by the mean and standard deviation of the predicted demand distribution. These serve as coarse-grained priors, establishing a baseline expectation of demand for short-term refinement.

The high-frequency features mentioned above as the fourth source include two types of demand deviation indicators, capturing fine-grained temporal dynamics.
The first type refers to intermediate features from the two-stage framework, capturing recent deviations between observed 15-minute demand and hourly predictions from stage 1. These indicators, reflecting short-term rental and return activity, are jointly used to inform both pickup and drop-off predictions.
The second type of deviation feature captures discrepancies in metro passenger flow. Specifically, it measures the difference between current check-ins or check-outs and their historical averages for the corresponding hour and weekday. For pickup demand forecasting, deviations in metro check-outs are included to reflect potential mode transfers from metro to bike-sharing. Conversely, deviations in metro check-ins are used for drop-off demand forecasting, capturing likely interchanges from bike-sharing to metro during commuting hours.





\subsection{Predictive Accuracy Evaluation Metrics and Key Performance Measures}
\label{subsect:predictive_accuracy_evaluation_metrics}
Predictive accuracy is evaluated using both deterministic and probabilistic metrics. These measures assess the model’s ability to provide accurate point forecasts as well as to capture uncertainty in the predicted distribution.  

Mean Absolute Error (MAE) and Root Mean Squared Error (RMSE) are used to evaluate the deterministic prediction accuracy. Given the group of actual values $y_i$ and their corresponding predicted values $\hat{y}_i$, the Mean Absolute Error (MAE) is defined as, 
\begin{equation}
    \text{MAE} = \frac{1}{N} \sum_{i=1}^{N} |y_i - \hat{y}_i|,
\end{equation}

where $N$ is the total number of data points.
And the Root Mean Squared Error (RMSE) is given by,
\begin{equation}
\text{RMSE} = \sqrt{\frac{1}{N} \sum_{i=1}^{N} (y_i - \hat{y}_i)^2}.
\end{equation}

Probabilistic predictors typically return the estimated parameters of the predicted demand distribution, which can be used to generate predictive samples. In our experiments, we generate 100 sample points for each forecast, and the point forecast is taken as the median of these sampled values.

For probabilistic demand forecasting performance, metrics Continuous Ranked Probability Score (CRPS) and Interval Score (IS) are applied. 
CRPS is a comprehensive measure that quantifies forecasting uncertainties by taking the full probabilistic distributions into account.
Based on Gneiting and Raftery's formulation \citep{gneiting2007strictly}, when the actual value $y$ is a scalar, the CRPS is defined as follows:
\begin{equation}
    \mathrm{CRPS}(F, y) = \int_{\mathbb{R}} \left[ F(\hat{y}) - \mathbf{1}_{\{\hat{y} \ge y\}} \right]^2 \, d\hat{y},
\end{equation}
In this formulation, the empirical CDF of the $F(\hat{y})$ is estimated by a step function $\mathbf{1}(\hat{y} \geq y)$, returning 1 when the predicted value is higher or equal to the actual observation, otherwise 0. The values of CRPS are always non-negative, ranging from 0 to $+ \infty$ by definition, with smaller values indicating better-calibrated probabilistic forecasts.

While CRPS measures the overall probabilistic difference between the actual and predicted distributions, the Interval Score (IS) focuses on the quality of specific prediction intervals. IS accounts for both the width of the interval and whether the actual observation falls within it. The calculation of IS follows:
\begin{equation}
    \text{IS}_{\alpha} = (\hat{u} - \hat{l}) + \frac{2}{\alpha} (\hat{l} - y) \mathbf{1}_{\{y < \hat{l}\}} + \frac{2}{\alpha} (y - \hat{u}) \mathbf{1}_{\{y > \hat{u}\}},
\end{equation}
where $[\hat{l}, \hat{u}]$ is the prediction interval associated with a nominal coverage probability denoted by $1 - \alpha$, and $y$ is the actual value. In this study, we set $\alpha$ to be 0.1, meaning the lower bound $l$ corresponds to the $5^{th}$ percentile of the predicted distribution, while the upper bound $\hat{u}$ corresponds to the $95^{th}$ percentile. A smaller IS value indicates a better quality of the generated prediction interval, which balances precision and reliability.

To summarize, both CRPS and IS evaluate how close the predicted distribution is to the actual value, while penalizing poorly calibrated probabilistic forecasts, such as under-confident predictions with wide confidence intervals.
While both CRPS and IS assess the quality of probabilistic forecasts at the individual instance level, we compute the Mean CRPS (MCRPS) and Mean IS (MIS) to summarize the overall average performance across a group, such as per station or within the same time period.


%

The demand forecasting task for an urban bike-sharing service involves predicting multiple time intervals into the future across hundreds of locations within the service network. 
To provide a thorough and practical assessment of forecasting performance, we adopt a multi-dimensional evaluation approach. Rather than relying solely on aggregate accuracy, we examine model performance across spatial and temporal contexts to better understand its strengths and limitations in real-world operations. This includes evaluating overall performance to measure general accuracy, analyzing spatial variations to assess consistency across stations with differing demand profiles, and exploring temporal dynamics to detect fluctuations in predictive reliability over time. Such a comprehensive evaluation framework enables us to assess not only how well the model performs on average, but also its robustness across diverse scenarios (e,g., varying station characteristics, usage patterns, and operational conditions) that are critical for deploying predictive model in large-scale and dynamic bike-sharing systems.

\subsection{Benchmarking Design}
\label{subsect:benchmarking_design}
To validate the performance of the Transformer-based T-STAR framework in the multivariate short-term demand forecasting task for the Washington D.C. case study, we conduct two series of benchmarking experiments. These aim to evaluate the impact of contextual input design and compare T-STAR's performance against both classic and state-of-the-art forecasting models, offering insights into the most effective configurations for accurate and robust bike-sharing demand prediction.

\subsubsection{Series 1: Contextual Input Strategies}
\label{subsubsect:contextual_input}
The first set of benchmarking experiments evaluates how contextual information contributes to demand forecasting performance. Prior work suggests that contextual features can improve both accuracy and reliability, particularly under changing operating conditions and external disruptions. Accordingly, we test progressively richer contextual feature sets to quantify their marginal impact on forecasting accuracy.


\begin{table}[htbp]
\centering
\caption{Summary of input combinations evaluated in the Washington D.C. case study. \cmark indicates that the input was included in the model.}
\resizebox{\textwidth}{!}{%
\begin{tabular}{|r|c|c|c|c|c|}
\hline
\multirow{2}{*}{\textit{\textbf{Input Combination Types}}} & \multicolumn{4}{c|}{\textbf{Contextual Features}} & \textbf{Time Series Info.} \\
\cline{2-6}
& \makecell{Temporal variables \\(Hour-of-a-day and \\ Day-of-the Week)} 
& \makecell{External environment\\(Weather \& Holiday)} 
& \makecell{Bike-Sharing facility \\ (dock  capacity) and recent \\ deviations from connected \\ metro check-in/out flow} 
& \makecell{Past demand variations} 
& \makecell{Previous observations\\from the time series} \\
\hline
\textit{Baseline} & \cmark &  &  &  & \cmark \\
\textit{Global Environment} & \cmark & \cmark &  &  & \cmark \\
\textit{Environment, Facility and PT} & \cmark & \cmark & \cmark &  & \cmark \\
\makecell[r]{\textit{\textbf{T-STAR}}} & \cmark & \cmark & \cmark & \cmark & \cmark \\
\hline
\end{tabular}%
}
\label{tab:input_features_summary}
\end{table}
The features explored in this our experiments are categorized into four input combinations:
\begin{itemize}
    \item \textit{\textbf{Baseline}}: This basic category only considers temporal variables, hour-of-the-day and day-of-the-week, to capture typical cyclical patterns in demand. These features are derived directly from the datetime information of the demand time series and represent internal temporal structure of the past and future values on the time series.
    \item \textit{\textbf{Global Environment}}: In addition to temporal features, this category includes contextual factors such as weather conditions and public holidays, which influence overall mobility patterns across the city. These features are shared across all stations and represent conditions that affect bike-sharing demand from the environment external to the bike-sharing operations.
    \item \textit{\textbf{Environment, Facility and PT}}: This advanced category extends the \textit{Global Environment} setting by incorporating station-level dock capacity, which may influence future demand, and recent deviations in check-in and check-out passenger flows at nearby metro stations. Together, these features aim to capture the effects of local infrastructure and public transport dynamics on short-term bike-sharing usage.
    \item \textit{\textbf{T-STAR}}: The proposed T-STAR framework adopts a comprehensive combination of inputs. It integrates all previous contextual features and additionally includes recent demand deviation features, matched precisely to hourly and short-term predictions based on these features’ temporal granularity. These internal demand deviation features are calculated by comparing recent short-term observed demand against corresponding hourly demand forecasts generated by the first-stage long-term predictive model. The objective is to improve forecasting accuracy by explicitly capturing recent short-term demand fluctuations.
\end{itemize}


\subsubsection{Series 2: Base Predictors}
\label{subsect:base_predictors}

The second series of our benchmarking exercise investigates the performance of different base predictive models, each capturing distinct forecasting assumptions and strengths. This exploration helps identify the effectiveness and appropriateness of various forecasting methods in the context of short-term bike-sharing demand prediction. The selected base predictors are introduced in the following.

\paragraph{Historical Average}
This straightforward and deterministic baseline method forecasts future demand as the average of historical observations from the same day-of-the-week and hour-of-the-day, leveraging the recurring seasonal patterns of demand.
\paragraph{Myopic Approach}
This approach utilizes the immediate past demand observation as the forecast for the subsequent time interval. This simplistic predictor serves as a practical baseline for short-term forecasting scenarios, under the assumption of minimal demand fluctuations between consecutive intervals.
\paragraph{Time Series Transformer (TST) - Hourly Model}
This approach generates 15-minute demand forecasts by proportionally dividing hourly predictions produced by the TST model used in Stage 1 of the T-STAR framework. The hourly predictions leverage past hourly demand, temporal factors (day-of-the-week and hour), and weather information. By operating at a coarser resolution, the model helps mitigate the sparsity and noise often present in high-frequency data, resulting in more stable predictions. However, unlike the T-STAR framework, this approach does not incorporate high-resolution temporal signals, limiting its responsiveness to short-term demand fluctuations.
\paragraph{XGBoost (Extreme Gradient Boosting)}
Introduced by \cite{chen2016xgboost}, XGBoost is a machine learning technique renowned for its predictive performance and computational efficiency. Its effectiveness in short-term transportation demand forecasting arises from robustness against sparse data, efficient handling of nonlinear relationships, and resistance to overfitting. However, it is limited by its reliance on historical examples, leading to weaker performance when predicting scenarios unseen during training. Additionally, XGBoost's boosting structure is inherently designed for point prediction and does not support probabilistic forecasting by default.

In addition, we look into the application of six state-of-the-art neural-network predictors, each described below.
\paragraph{STGCN and STGCN-VAE}

Originally introduced by \cite{yu2017spatio}, the Spatio-Temporal Graph Convolutional Network (STGCN) addresses forecasting challenges in networked systems by integrating spatial graph convolutions with gated 1D temporal convolutions. This fully convolutional architecture enables the scalable learning of complex dynamics while explicitly leveraging the underlying network topology. Consistent with our Transformer-based predictors, we provide station-specific attributes as node features to learn representative station embeddings. To enable probabilistic forecasting for the standard STGCN, we extend the architecture with a distributional output head and train the model by minimizing the negative log-likelihood.
In this study, we also include STGCN-VAE as a state-of-the-art graph-based benchmark for nonparametric probabilistic prediction \citep{peng2025uncertainty}. This model utilizes an STGCN backbone to extract spatio-temporal features, which are subsequently compressed into a latent space via a Variational Autoencoder (VAE) and processed through a Kernel Density Estimation (KDE) module to construct predictive distributions.
For each graph-based predictor, we evaluated three commonly applied graph construction strategies to define the static adjacency matrix. To ensure an optimal baseline, we report the results corresponding to the best-performing configuration for each benchmark. Technical details regarding these construction methods and the final configurations selected for each model are provided in \ref{appendix:graph_construction}.
\paragraph{DeepAR} Introduced by \cite{salinas2020deepar}, DeepAR is a probabilistic autoregressive forecasting model utilizing LSTM architecture. It excels at capturing temporal dependencies within time series data and serves as a global predictor for all bike-sharing stations. However, due to its sequential nature, it can struggle with sparse or intermittent demand series, resulting in potential instability and inaccuracies when forecasting infrequent or zero-inflated demand.
\paragraph{\textcolor{black}{STAEformer}} \textcolor{black}{Spatio-Temporal Adaptive Embedding Transformer (STAEformer) was introduced by \cite{liu2023staeformer}. It augments a vanilla Transformer encoder with spatio-temporal adaptive embeddings that tie node representations to spatio-temporal attention patterns, thereby injecting spatial and temporal inductive bias and improving forecasting performance. Following the original design, STAEformer uses historical time-series from all locations as input, together with hour-of-day and day-of-week features as contextual attributes \citep{liu2023staeformer}. For station-wise 15-minute bike-sharing demand forecasting, we adopt the same input structure and further extend STAEformer with a distributional output head. The model is trained by minimizing the negative log-likelihood to produce probabilistic demand forecasts.}
\paragraph{Temporal Fusion Transformer (TFT)} Recently proposed by \cite{lim2021temporal}, TFT combines LSTM with a self-attention mechanism, effectively capturing joint temporal and contextual relationships. It incorporates a gating mechanism for adaptive feature selection, reducing the risk of overfitting despite numerous input features. Its complexity, however, can lead to extensive computational demands during training. Different from DeepAR and TST, TFT adopts pinball loss as training objective. 
\paragraph{Simple Time Series Transformer (Simple TST)} 
As introduced in Section \ref{sect:transformer}, TST is used as the base predictor in the proposed T-STAR forecasting framework. To assess the added value of T-STAR’s two-stage architecture, we implement a one-stage baseline model, referred to as \textit{Simple TST}. This processes all available contextual information in a single stage to directly generate per-15-minute short-term demand forecasts. 

\textcolor{black}{We tuned model hyperparameters using a five-fold time-series cross-validation scheme that preserves temporal order and avoids information leakage. Specifically, we employed a rolling forecast origin design with an expanding training window and a fixed one-week validation window per fold. The five validation weeks were distributed across the three-month training period to ensure coverage of weekday/weekend demand patterns and to capture atypical demand intervals when present. Hyperparameter configurations were selected by minimizing the average validation RMSE across the five folds. For each model, 30 trials (i.e., 30 different hyperparameter combinations) were performed with Optuna package to efficiently explore the hyperparameter search space \citep{akiba2019optuna}.}
The types and search ranges of the tuned hyperparameters are detailed in the \ref{appendix:hyperparameter}, along with the selected values for each model.



\section{Results and Discussion}
\label{chapter:results_and_discussion}

\subsection{Short-Term Demand Forecasting Performance}
\label{subsect:short-term_demand_forecasting_performance}

\subsubsection{Overall Predictive Accuracy and Computational Efficiency}
\label{subsubsect:overall_performance}

This subsection reports the overall forecasting performance of selected predictors and corresponding analysis. To account for variability due to random initialization, we performed multi-seed training for each predictor. Specifically, we trained five independent model instances using five randomly selected seeds, while keeping the hyperparameter configuration and experimental protocol identical. The results reported in this section are aggregated across seeds by taking the average across runs.

We start with analyzing the overall predictive accuracy and computational efficiency of the proposed Transformer-based T-STAR framework for short-term bike-sharing demand forecasting, in comparison to selected benchmark methods.
Each forecasting approach consists of two key components: the base predictive model and the contextual input features incorporated to support forecasting. We evaluate eight different forecasting approaches, including methods that produce only deterministic forecasts as well as those capable of generating both deterministic and probabilistic outputs.
To provide a generalized assessment of model performance, we report the aggregated mean accuracy across all stations. In addition, we examine variations in predictive performance across different stations and over time to evaluate the robustness of each forecasting approach. This analysis allows us to assess whether a single global model can maintain consistent predictive accuracy under diverse spatial and temporal conditions.


Table \ref{tab:overall_pickup} summarizes the overall predictive performance of the selected forecasting approaches for pickup demand, evaluated over a 20-day period across 235 Capital Bikeshare stations in Washington, D.C.
The Transformer-based T-STAR framework consistently delivers accurate and robust performance across all key metrics, including deterministic accuracy (measured by MAE and RMSE) and probabilistic accuracy (measured by MCRPS and MIS). Among the selected SOTA probabilistic models, Transformer-based T-STAR achieves the lowest MAE, RMSE, and MCRPS.
Compared to the historical average predictor, Transformer-based T-STAR reduces the MAE by 54.9\% and the RMSE by 7.9\%. Furthermore, relative to a Transformer predictor without contextual inputs, Transformer-based T-STAR improves performance by 19.7\%, 7.4\%, 27.7\%, and 50.3\% in MAE, RMSE, MCRPS, and MIS, respectively. These large improvements demonstrate the substantial benefits of two-stage adaptive contextual representation.

Although XGBoost and the hourly Transformer predictor achieve the lowest RMSE, this apparent advantage is partly driven by the demand series’ pronounced zero inflation and heavy-tailed spikes, under which RMSE disproportionately rewards strategies that avoid large squared errors.
In such cases, XGBoost tends to bias its predictions toward occasional high-demand instances in order to minimize the training loss, at the expense of accuracy on the more frequent low- or zero-demand observations. As a result, the forecasting experiment showed that XGBoost tends to over-predict during off-peak periods and underperforms at low-demand stations. The quarterly forecasts derived from the hourly Transformer benefit from temporal aggregation. The hourly forecasts tend to be smoother and less volatile, so distributing them across four 15-minute intervals dampens extremes and improves RMSE. However, this uniform within-hour allocation cannot capture sub-hour demand bursts driven by short-term factors, yielding persistent interval-level deviations and therefore higher MAE than multivariate predictors that leverage short-term contextual inputs.

Both the historical average and myopic predictors exhibit significantly larger errors across all evaluation metrics. 
Moreover, STAEformer’s spatio-temporal adaptive embeddings and attention improve point forecasting by learning flexible inter-station and temporal dependencies from historical demand, which aligns with its relatively low MAE/RMSE. However, with inputs largely limited to past demand and coarse calendar features, this method has less information to condition the uncertainties on exogenous drivers of dispersion and rare spikes. Context-rich probabilistic models (DeepAR, TFT, and Transformer-based T-STAR) therefore deliver better-calibrated and sharper predictive distributions, as evidenced by their lower MCRPS and MIS.
This finding highlights the importance of incorporating contextual information and capturing dynamic temporal trends for adaptive forecasting and uncertainty quantification, compared to relying solely on the general seasonal demand patterns or the most recent observations. 

Although finer temporal granularity increases challenges due to greater demand sparsity and variability, the short-term Transformer-based T-STAR forecasting approach maintains high predictive accuracy and performance robustness at this resolution. These improvements justify the efforts associated with T-STAR's two-stage training structure to enable more precise and timely operational decision-making in bike-sharing systems.

\begin{table}[h!]
\centering
\caption{Predictive accuracy comparison of short-term pickup demand forecasting models across different approaches. The aggregated mean accuracy across all stations is reported before the parentheses. Metric values are averaged over five runs with different random seeds for each predictive model (except the Historical Average and Myopic approaches). The first value in parentheses indicates the standard deviation across stations, and the second value indicates the standard deviation across time intervals (aggregated across stations at each interval). Probabilistic metrics (MCRPS and MIS) are only reported for models supporting probabilistic forecasting.}
\resizebox{\textwidth}{!}{%
\begin{tabular}{ccc>{\color{black}}c>{\color{black}}c>{\color{black}}c>{\color{black}}c}
\hline
\textbf{\makecell{Forecasting \\ Type}} &\textbf{Basic Predictor} & \textbf{Input Types} & \textbf{MAE} & \textbf{RMSE} & \textbf{MCRPS} & \textbf{MIS} \\
\hline
&\textit{Historical Average} & \textit{Baseline} & 0.335 (0.150; 0.219) & 0.504 (0.187; 0.260) & -- & -- \\
\multirow{2}{*}{Deterministic} &\textit{Myopic} & \textit{\makecell{Last demand observa-\\tion of each station}} & 
0.229 (0.115; 0.200) & 0.578 (0.180; 0.312) & -- & -- \\
&\textit{TST-Hourly} & \textit{Global Environment} & 0.204 (0.111; 0.165) & \textbf{0.436 (0.138; 0.235)} & -- & -- \\
&\textit{XGBoost} & \textit{Env., Facilty and PT} & 0.334 (0.068; 0.120) & 0.459 (0.133; 0.205) & -- & -- \\
\hline
&\textit{Simple TST} & \textit{Baseline} & 
0.188 (0.123; 0.176) & 0.501 (0.197; 0.310) & 0.177 (0.109; 0.156) & 3.753 (1.276; 2.750) \\
& \textit{STAEformer} & \textit{Baseline} &
0.152 (0.096; 0.154) & 0.475 (0.180; 0.289) & 0.187 (0.054; 0.084) & 3.628 (0.206; 0.286) \\
\multirow{3}{*}{Probabilistic} &\textit{DeepAR} & \textit{Env., Facilty and PT} & 
0.159 (0.097; 0.156) & 0.474 (0.169; 0.280) & 0.133 (0.071; 0.119) & 2.188 (0.756; 1.616) \\
&\textit{STGCN} & \textit{Env., Facilty and PT}  &
0.422 (0.049; 0.332) & 0.713 (0.071; 0.342) & 0.399 (0.029; 0.200) & 7.465 (0.446; 1.277) \\
& \textit{STGCN-VAE} & \textit{Env., Facility and PT} &
0.404 (0.057; 0.087) & 0.500 (0.124; 0.175) & 0.398 (0.057; 0.087) & 15.419 (2.286; 3.463) \\
&\textit{TFT} & \textit{Env., Facilty and PT} & 
0.156 (0.095; 0.155) & 0.473 (0.171; 0.281) & 0.134 (0.067; 0.115) & 2.123 (0.653; 1.142) \\
&\textit{\makecell{Transformer-Based \\T-STAR}}& \textit{T-STAR} & 
\textbf{0.151 (0.089; 0.148)} & 0.464 (0.156; 0.271) & \textbf{0.128 (0.062; 0.105)} & \textbf{1.864 (0.508; 0.925)} \\
\hline
\multicolumn{7}{l}{\makecell[l]{(1) Since the \textit{Historical Average}, \textit{Myopic}, \textit{TST-Hourly}, and \textit{XGBoost} models do not produce probabilistic forecasts in our experiments, the probabilistic \\ evaluation metrics \textbf{MCRPS} and \textbf{MIS} are not reported for these models (denoted by -- in the table).}}\\
\multicolumn{7}{l}{\makecell[l]{(2) The best-performing values for each metric are highlighted in bold in the table.}}\\
\end{tabular}%
}
\label{tab:overall_pickup}
\end{table}


Table \ref{tab:overall_dropoff} presents the overall predictive performance of the selected forecasting approaches for short-term drop-off demand.
Consistent with the results for pickup demand, the Transformer-based T-STAR framework demonstrates strong performance on the drop-off task, highlighting its ability to generalize predictive quality to a different but closely related demand series.
Transformer-based T-STAR achieves the lowest MAE (0.151), MIS (1.895), and ties for the lowest MCRPS (0.126), indicating solid performance in both point and distributional predictions. 

Interestingly, most models exhibit slightly better accuracy for drop-off than for pickup demand. This trend suggests that drop-off patterns may be inherently easier to forecast at a 15-minute resolution, potentially because drop-offs are less constrained by facility availability compared to pickups. For the Capital Bikeshare service, riders can opt to pay a small surcharge to park at public bike racks if docking stations are full. The flexibility in parking reduces the sensitivity of drop-off behavior to real-time bike-sharing station capacity (in parking docks). However, pickup is hard constrained by the fleet availability at the origin station.

As showed in table  \ref{tab:overall_dropoff}, for drop-off demand, the vanilla Transformer without contextual inputs performs comparably to the T-STAR Transformer across MAE, RMSE, and MCRPS. This indicates that contextual information contributes less to forecasting short-term drop-offs than it does for pickups. 
However, T-STAR yields a markedly lower MIS. This indicates that T-STAR’s two-stage design, together with short-term variation signals and station-specific attributes, improves uncertainty quantification by producing tighter and better-calibrated 95\% prediction intervals, thereby reducing out-of-bound interval penalties on unseen data. We also observe the lowest RMSE for XGBoost and the hourly TST variant, likely due to temporal smoothing from their modeling and/or aggregation choices.
Overall, results from the Capital Bikeshare case study indicate that drop-off dynamics are comparatively less sensitive to contextual variations, and that a context-free Transformer is sufficient for capturing short-term fluctuations, while T-STAR primarily adds value through improved interval reliability.

\begin{table}[h!]
\centering
\caption{Predictive accuracy comparison of short-term drop-off demand forecasting models across approaches. Metric values are averaged over five runs with different random seeds for each predictive model (except the Historical Average and Myopic approaches). The aggregated mean accuracy across all stations is reported before the parentheses. The first value in parentheses indicates the standard deviation across stations, and the second value indicates the standard deviation across time intervals (aggregated across stations at each interval). Probabilistic metrics (MCRPS and MIS) are only reported for models supporting probabilistic forecasting.}
\resizebox{\textwidth}{!}{%
\begin{tabular}{ccc>{\color{black}}c>{\color{black}}c>{\color{black}}c>{\color{black}}c}
\hline
\textbf{\makecell{Forecasting \\ Type}} & \textbf{Basic Predictor} & \textbf{Input Types} & \textbf{MAE} & \textbf{RMSE} & \textbf{MCRPS} & \textbf{MIS} \\
\hline
& \textit{Historical Average} & \textit{Baseline} & 
0.331 (0.153; 0.220) & 0.505 (0.199; 0.270) & -- & -- \\
\multirow{2}{*}{Deterministic} 
& \textit{Myopic} & \textit{\makecell{Last demand observa-\\tion of each station}} & 
0.225 (0.114; 0.199) & 0.572 (0.187; 0.316) & -- & -- \\
& \textit{TST-Hourly} & \textit{Global Environment} & 
0.199 (0.109; 0.165) & \textbf{0.437 (0.146; 0.242)} & -- & -- \\
& \textit{XGBoost} & \textit{Env., Facility and PT} & 
0.317 (0.071; 0.126) & 0.452 (0.140; 0.217) & -- & -- \\
\hline
& \textit{Simple TST} & \textit{Baseline} & 
\textbf{0.151 (0.091; 0.149)} & 0.462 (0.165; 0.276) & \textbf{0.126 (0.065; 0.111)} & 2.202 (0.645; 1.446) \\
& \textit{STAEformer} & \textit{Baseline} & 
0.152 (0.097; 0.156) & 0.475 (0.190; 0.297) & 0.186 (0.055; 0.087) & 3.655 (0.227; 0.345)\\
\multirow{3}{*}{Probabilistic} 
& \textit{DeepAR} & \textit{Env., Facility and PT} & 
0.155 (0.096; 0.155) & 0.473 (0.181; 0.288) & 0.133 (0.069; 0.116) & 2.011 (0.743; 1.338) \\
& \textit{STGCN} & \textit{Env., Facility and PT} &
0.602 (0.066; 0.479) & 0.917 (0.069; 0.461) & 0.535 (0.036; 0.283) & 8.969 (0.201; 1.946) \\
& \textit{STGCN-VAE} & \textit{Env., Facility and PT} &
0.395 (0.059; 0.092) & 0.498 (0.132; 0.188) & 0.390 (0.059; 0.092) & 15.171 (2.369; 3.662) \\
& \textit{TFT} & \textit{Env., Facility and PT} & 
0.154 (0.095; 0.156) & 0.472 (0.181; 0.290) & 0.131 (0.068; 0.117) & 2.083 (0.659; 1.223) \\
& \textit{\makecell{Transformer-Based \\T-STAR}} & \textit{T-STAR} & 
\textbf{0.151 (0.109; 0.216)} & 0.462 (0.173; 0.284) & \textbf{0.126 (0.078; 0.160)} & \textbf{1.895 (0.973; 2.334)}\\
\hline
\multicolumn{7}{l}{\makecell[l]{(1) Since the \textit{Historical Average}, \textit{Myopic}, \textit{TST-Hourly}, and \textit{XGBoost} models do not produce probabilistic forecasts in our experiments, the probabilistic \\ evaluation metrics \textbf{MCRPS} and \textbf{MIS} are not reported for these models (denoted by -- in the table).}}\\
\multicolumn{7}{l}{\makecell[l]{(2) The best-performing values for each metric are highlighted in bold in the table.}} \\

\end{tabular}%
}
\label{tab:overall_dropoff}
\end{table}

For both pickup and drop-off forecasting, graph-based predictors STGCN and STGCN-VAE are consistently outperformed by the T-STAR Transformer and other neural-network-based predictors. Their sub-optimal performance likely stems from a weak spatial signal relative to the dominant, station-specific temporal dynamics present at a 15-minute granularity.
Notably, our experiments demonstrate that while STGCN-VAE improves point-prediction metrics over the original STGCN, it exhibits significantly poorer predictive interval calibration. Although STGCN-VAE’s nonparametric approach offers flexibility by relaxing rigid distributional assumptions, our results suggest its efficacy is highly sensitive to data granularity. In our high-resolution (15-minute, station-level) setting, STGCN-VAE yielded the highest MIS among all benchmarks, nearly doubling the next closest baseline. This diverges from the findings of \cite{peng2025uncertainty}, where STGCN-VAE excelled in calibration for bike-sharing demand data aggregated at a coarser spatial (postal-code regional) and temporal (hourly) level. In our case study, the shift to high-granularity time series introduces extreme sparsity and increases stochasticity, challenging the VAE’s ability to reconstruct reliable latent distributions. Ultimately, these results underscore that the performance of nonparametric generative models can be contingent upon the spatio-temporal scale, highlighting the importance of scale-aware model selection for high-resolution urban mobility operations.

To further investigate the performance gap of graph-based predictors, we conducted a partial correlation analysis to isolate spatial dependencies from shared temporal patterns, for both pickup and drop-off demand time series. Specifically, we compared the station-wise partial correlations obtained for raw demand time series with those for demand residuals. The residual latter obtained by removing seasonal averages (grouped by day-of-week and hour) to isolate stochastic, high-frequency fluctuations.
Our analysis show that, while raw data exhibits a modest average correlation ($\approx$ 0.139) due to shared daily cycles, this correlation is reduced by over 68\% (to $\approx$ 0.044) once seasonal effects are removed. Analysis of the demand residuals shows that the most correlated 5\% of station pairs reach a maximum of only 0.076, confirming that at 15-minute intervals, station-wise demand shifts are nearly independent across the network. This empirical lack of spatial autocorrelation suggests that at high temporal resolutions, any underlying spatial dependency is superseded by stochastic noise. Consequently, graph-based approaches fail to extract a coherent spatial signal, whereas temporal-focus predictors excel by concentrating on the more reliable, station-specific contextual patterns.

In order to evaluate the statistical significance of the observed performance gains, we conducted a station-wise Wilcoxon signed-rank test comparing T-STAR transformer against all baseline models. For each station, error metrics were averaged across five independent experimental seeds to mitigate stochastic training effects ($N=235$). This two-sided paired test was employed to determine whether the distribution of station-wise metric values for T-STAR differed significantly from the benchmarks at a 95\% confidence level. The test results for both pickup and drop-off demand consistently yielded $p$-values below 0.05 across all comparisons. These findings demonstrate that the improvements achieved by T-STAR are statistically significant and spatially consistent.

To assess robustness, we quantified run-to-run variability across five independent random seeds using the coefficient of variation (CV). Most architectures showed relative variability below 5\%, whereas STGCN displayed markedly higher instability, with CV exceeding 10\% on multiple metrics. Overall, this is consistent with greater sensitivity of fixed-adjacency graph–temporal convolutional models to sparse and intermittent demand, where the adjacency-induced inductive bias and convolutional parameterization may interact with stochastic optimization. In contrast, the attention-based baselines were comparatively more stable across seeds in our experiments.

\subsubsection*{Computational Efficiency}
From the perspective of computational efficiency, the Transformer-based T-STAR approach remains competitive relative to other baselines considered in this study. All selected predictors, except for Seasonal ARIMA, are able to generate forecasts for the next 15-minute interval within a few seconds.

In terms of training time, all deep learning models in our experiment were trained for 100 epochs using mini-batching with a batch size of 256. On a CPU-based computing environment, the average training times were approximately 5.5 minutes for Transformer, 6.5 minutes for DeepAR, and 11 minutes for TFT. Since the Transformer-based T-STAR framework consists of two sequential stages, each with its own Transformer model, the total training time is approximately 11 minutes. Training STGCN and STAEformer is more computationally demanding. On an NVIDIA A100 GPU, STGCN and STGCN-VAE requires around 6.5 minutes to train, while STAEformer requires 5.5 minutes.

For XGBoost, training a global model capable of making predictions across all stations increases data volume substantially. Although XGBoost does not require iterative training like deep learning models, it typically consumes a large amount of RAM. Nevertheless, model fitting was completed in approximately 2 minutes.

The historical average and myopic benchmark models are computationally lightweight, as they require minimal input features and no model training. In contrast, the standard Seasonal ARIMA model proved computationally infeasible in our experimentation. Since bike-sharing services operate continuously, modeling both hourly and weekly seasonal patterns requires a minimum seasonal period of 168. This long seasonal cycle significantly increases model complexity. Under these conditions, Seasonal ARIMA failed to converge within four hours of computation, and was therefore excluded from further analysis.

\subsubsection*{Predictive Performance Robustness of Transformer-Based T-STAR}

To assess the predictive stability and robustness of the T-STAR Transformer against varying data availability and temporal demand shifts, we conducted two time-preserving evaluations: a rolling forecast origin (expanding-window) experiment and sliding window cross-validation. The former simulates a real-world deployment pipeline with periodic retraining, while the latter evaluates model stability under short-term distribution shifts across different historical contexts. Detailed experimental setups and result tables are provided in \ref{appendix:robustness}.
The results indicate that T-STAR consistently outperforms the benchmark contextual TST (with Environment, Facility and PT inputs) across all experimental folds for both pickup and drop-off forecasting. Notably, T-STAR maintains superior performance even with limited training data in early rolling folds and exhibits lower station-wise variability when facing temporal shifts in demand patterns. These results confirm that T-STAR’s spatiotemporal-aware architecture enhances generalizability and robustness against the inherent volatility of urban shared micro-mobility demand.

\subsubsection{Station-Level Forecasting Accuracy}
\label{subsubsect:station_level_performance}
\begin{figure}[h!]
    \centering
    \includegraphics[width=0.72\linewidth]{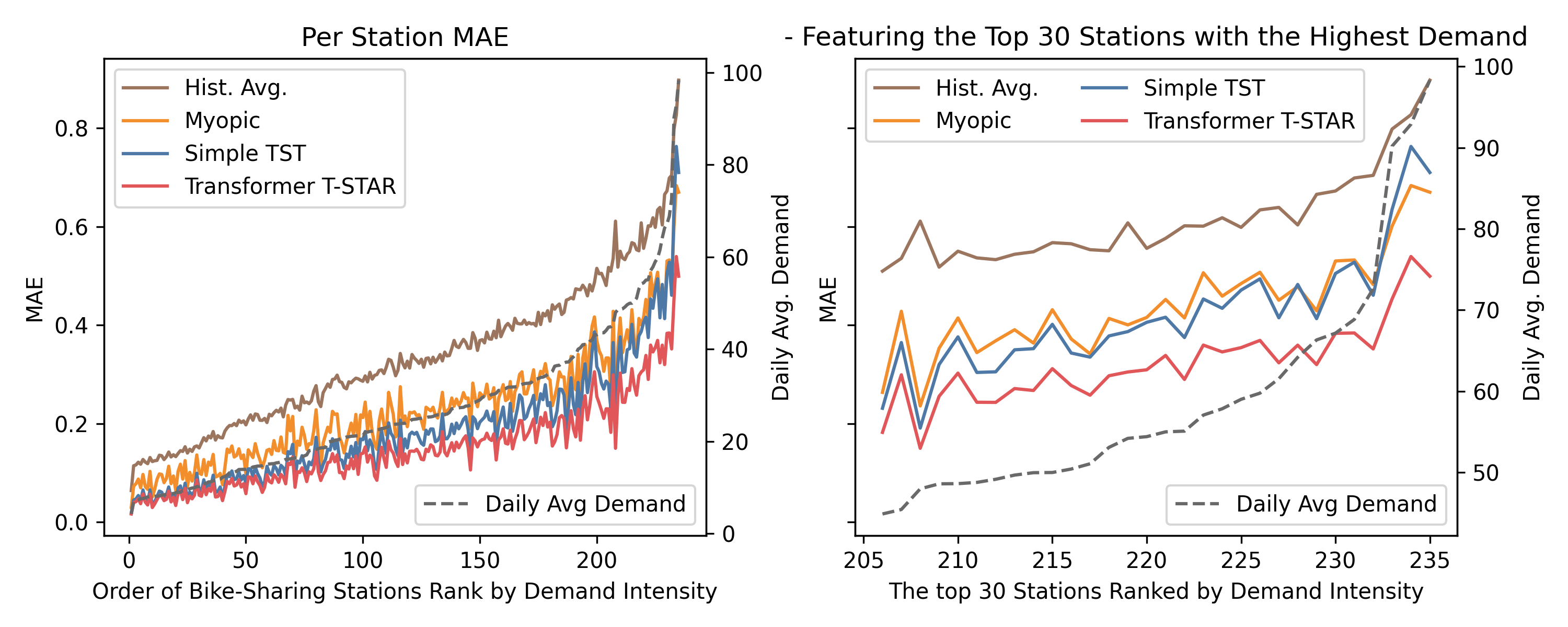}
    \caption{Station-level MAE by demand intensity in short-term pickup forecasting experiment. The left axis shows the MAE for each bike-sharing station, while the right axis indicates the corresponding daily average pickup demand. Stations along the horizontal axis are ordered by increasing average daily pickup demand. The left subplot presents MAE values for all 235 stations, and the right subplot provides a focused view of the top 30 highest-demand stations.The y-axis scales for both the left and right plots are matched to facilitate direct comparison.}
    \label{fig:station_MAE}
\end{figure}

\begin{figure}[h!]
    \centering
    \includegraphics[width=0.72\linewidth]{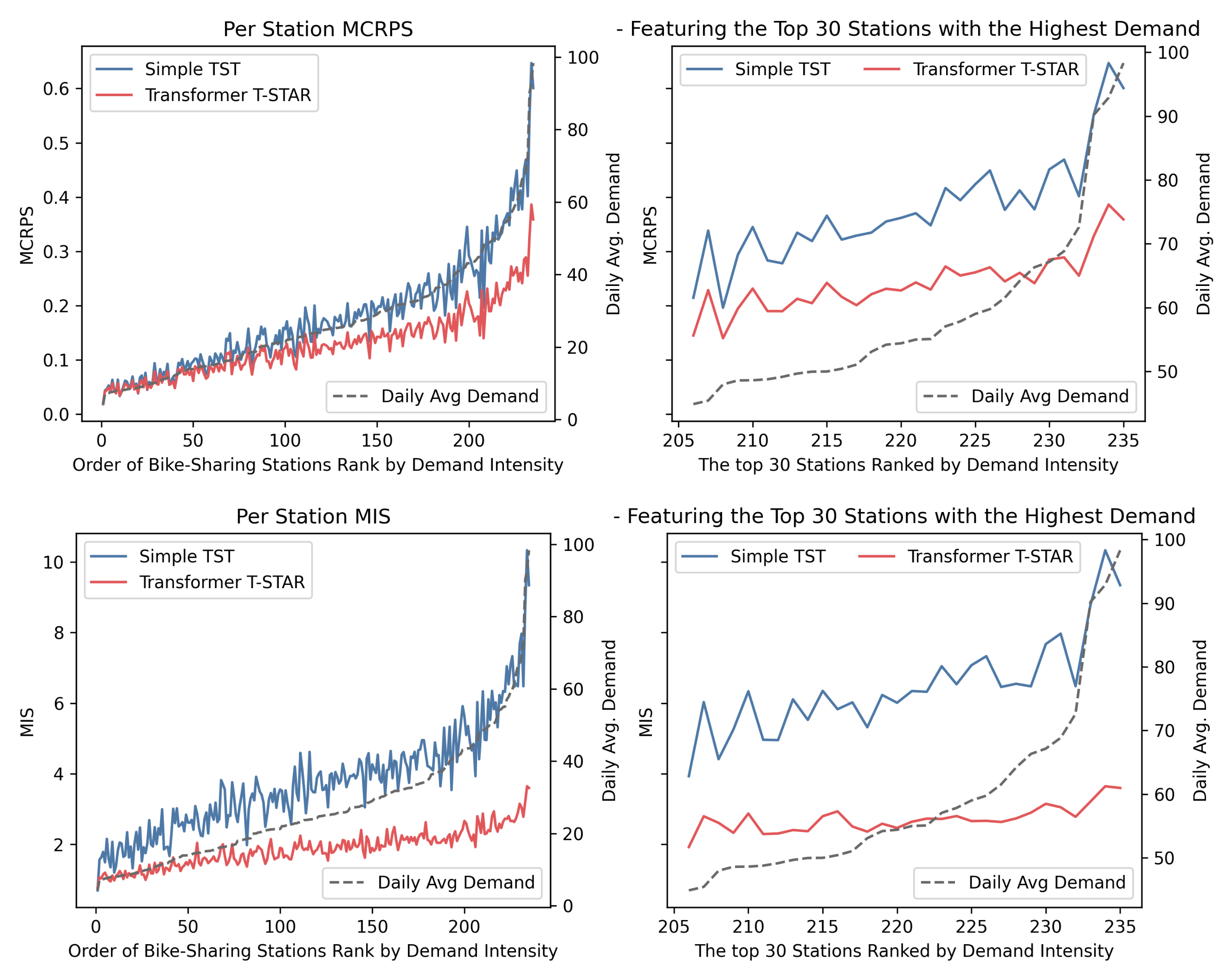}
    \caption{\textcolor{black}{Station-level MCRPS and MIS by demand intensity in short-term pickup forecasting experiment. The left axis shows the MCRPS and MIS for each bike-sharing station, while the right axis indicates the corresponding daily average pickup demand. Stations along the horizontal axis are ordered by increasing average daily pickup demand. The subplots on the left column present MCRPS and MIS values for all 235 stations, and the subplots from the right column provide a focused view of the top 30 highest-demand stations. The y-axis scales for both the left and right plots are matched to facilitate direct comparison.}}
    \label{fig:station_MIS}
\end{figure}

We provide a focused analysis of station-level forecasting performance by presenting the deterministic and probabilistic accuracy of 15-minute pickup demand predictions across all 235 bike-sharing stations in Figures \ref{fig:station_MAE} and \ref{fig:station_MIS}. The stations are ranked by their average daily demand, allowing us to examine how performance varies with demand intensity. A clear trend emerges, as both MAE and MIS generally increase with demand, indicating that short-term demand at high-activity stations is more challenging to predict accurately. This performance degradation is especially evident among the top five highest-demand stations. As shown in Figure \ref{fig:station_MAE}, the Transformer-based T-STAR model consistently outperforms baseline models, with the most substantial accuracy gains observed at high-demand stations. Similarly, Figure \ref{fig:station_MIS} demonstrates significant improvements in probabilistic forecasting accuracy when comparing the T-STAR model to a basic Transformer, further confirming the benefit of the two-stage contextual approach.

These results suggest that as demand levels increase, short-term demand patterns become more complex, likely due to a broader set of interacting factors related to system dynamics and surrounding facilities. The multivariate contextual representations learned by the T-STAR model are critical for capturing these fluctuations and associated uncertainties. The model’s forecasting advantage is especially pronounced at high demand nodes within the bike-sharing network, which typically play a central role in system operations. Improving accuracy at these high-impact locations can support more effective fleet management and service planning for micro-mobility services. 





\subsubsection{Forecasting Stability Across Time}
\label{subsubsect:temporal_performance}

The temporal accuracy analysis is included to assess the consistency of model performance throughout the testing period, helping to detect potential drift, degradation, and identify challenging periods to predict. In this subsection, we first examine the predictive performance of selected models across each 15-minute interval during the testing period for the short-term pickup demand forecasting task. 
Next, to evaluate model performance under rare or extreme conditions, we conduct separate analyses for normal and abnormal demand observations at the 15-minute level.

\begin{figure}[h!]
    \centering
    \includegraphics[width=0.85\linewidth]{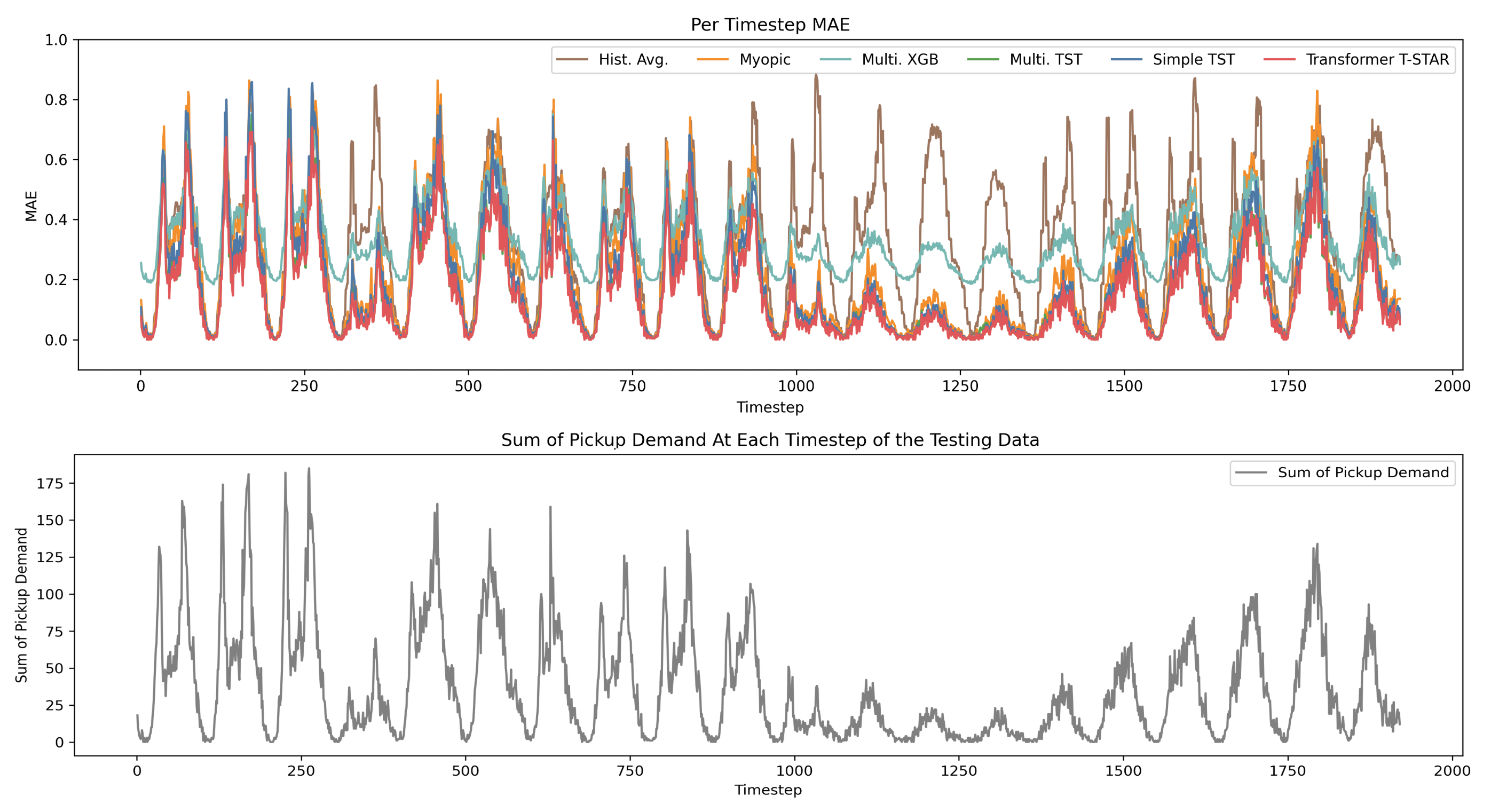}
    \caption{\textcolor{black}{Temporal dynamics of forecasting errors and actual pickup demand over the testing period. The top panel shows the per-timestep MAE of different forecasting models for short-term pickup demand. The bottom panel displays the corresponding total pickup demand at each 15-minute interval.}}
    \label{fig:pickup_temporal_MAE}
\end{figure}

The top panel of Figure \ref{fig:pickup_temporal_MAE} shows the MAE values of different predictive models over time, where each point represents the forecasting error across all bike-sharing stations at a given 15-minute interval. The bottom panel displays the corresponding interval's total pickup demand across the city. 
As expected, models that incorporate recent observations (all except Historical Average) demonstrate greater adaptability, particularly during unexpected drops in demand throughout the testing period. 
Notably, although the multivariate XGBoost predictor can respond to sudden demand shifts (e.g., between time steps 960 and 1440), it still shows relatively larger errors during abrupt demand drops compared to parametric Transformer-based models. This is likely because such scenarios with unexpected demand shifts were under-represented or absent in the training data, as XGBoost relies heavily on patterns observed during training and struggles to generalize beyond them.
Among all evaluated models, the Transformer-based T-STAR consistently achieves the lowest MAE across time, with especially strong performance during peak demand intervals. These results highlight Transformer-based T-STAR’s ability to capture regular demand patterns while also demonstrating robustness and responsiveness in forecasting short-term bike demand under varying and dynamic temporal conditions. This performance is attributed to its adaptive forecasting mechanism, which jointly leverages recent high-frequency demand variations and longer-term trends derived from hourly demand expectations.

\begin{figure}[h!]
    \centering
    \includegraphics[width=\linewidth]{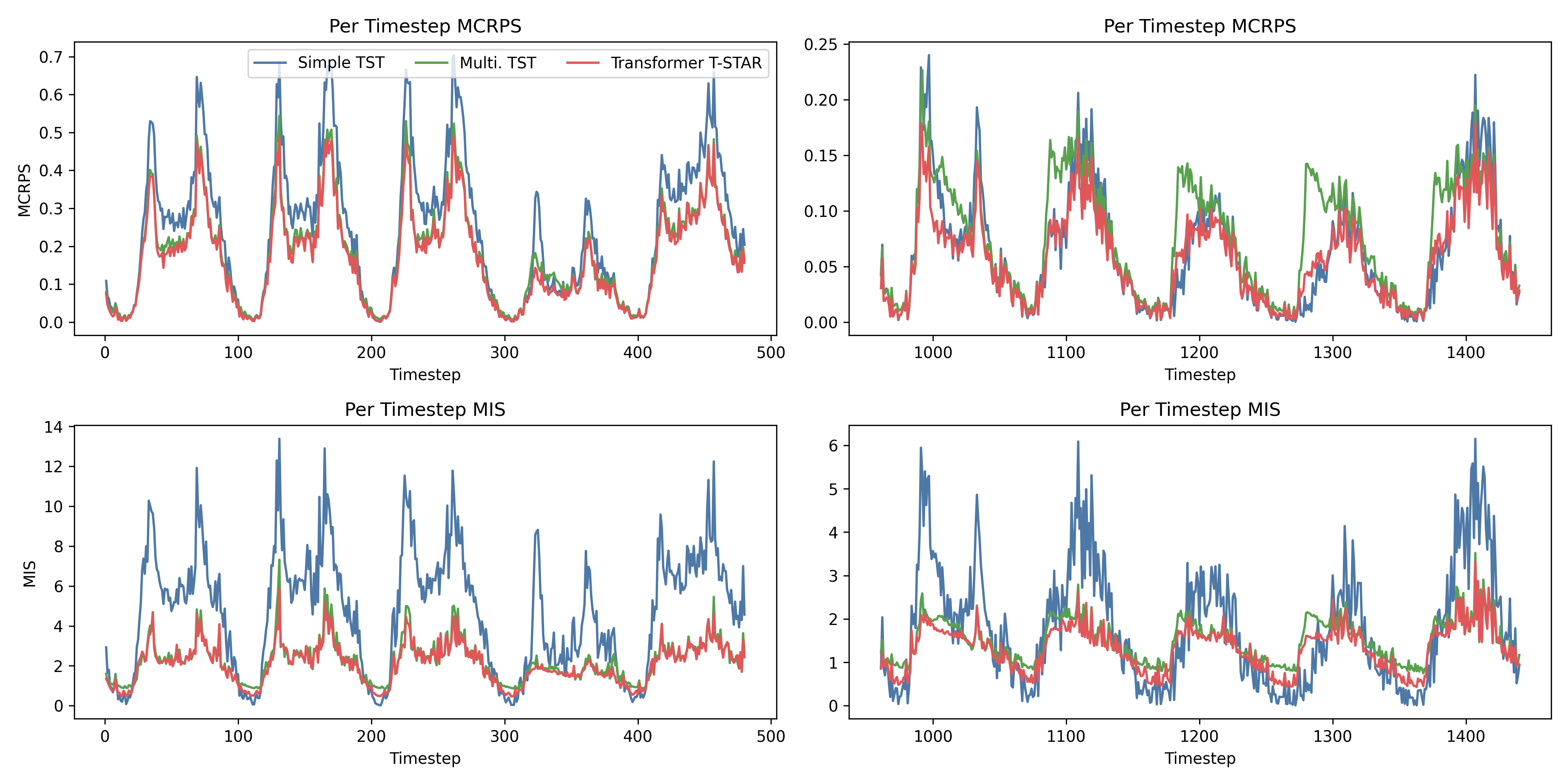}
    \caption{\textcolor{black}{Per-timestep comparison of probabilistic forecasting performance between the simple Transformer model (Simple TST), one-stage multivariate Transformer (Multi. TST) model with contextual features (Env., Facility and PT) and Transformer-based T-STAR. The left panels show performance over a regular demand period (Dec.11–16, 2022), while the right panels correspond to an unexpectedly low demand period (Dec.22–27, 2022).}}
    \label{fig:pickup_temporal_MCRPS_MIS}
\end{figure}

We further examine the impact of contextual inputs by comparing a period with regular demand patterns to one with an unexpectedly sharp demand drop, focusing on the probabilistic performance of one-stage Transformer models versus the Transformer-based T-STAR. The regular demand period is selected to be the first 5-day period from Dec.$11^{th}$ to Dec.$16^{th}$ (time steps 0-480), and the unexpectedly low demand period to be 5-day period from Dec.$22^{nd}$ to Dec.$27^{th}$ (time steps 960-1440) on the testing period. This drop coincided with a major Arctic blast that brought record-low temperatures and strong winds to Washington D.C., creating unfavorable conditions for outdoor biking. Additionally, the period included the three-day Christmas holiday, further reducing commuting demand in the city.

Figure \ref{fig:pickup_temporal_MCRPS_MIS} presents the per-timestep probabilistic forecasting performance of simple Transformer model, multivariate Transformer model with contextual features (Env., Facility and PT) and Transformer-based T-STAR, using MCRPS (top) and MIS (bottom) across two selected 5-day periods: a regular demand period (left) and an irregular low-demand period (right). Overall, T-STAR Transformer consistently outperforms the rest, achieving lower MCRPS and MIS values in both scenarios. The performance gain versus simple Transformer is especially evident during the regular demand period, suggesting the integration of short-term contextual embeddings effectively captures day-to-day demand fluctuations. During the irregular demand period with high uncertainty, all Transformer predictors recognize the dominant global influence (e.g., severe weather and holidays). However, the Transformer-based T-STAR variant consistently achieves better performance, as reflected by lower MCRPS and MIS. This pattern indicates that T-STAR’s two-stage contextual design more effectively integrates the compound effects of extreme weather and holidays, yielding forecasts that are both more accurate and more stable under regime shifts.

\subsubsection*{Abnormal and Normal Demand Prediction}
To evaluate model robustness under unusual scenarios, we conduct a targeted analysis comparing forecasting performance specifically on rare abnormal observations. In this experiment, we define abnormal observations as time steps where a station's demand has a z-score of 3 or higher, indicating a significant deviation from its historical average. These abnormal intervals account for approximately 1\% of the dataset. For this analysis, we focus exclusively on models using the TST as the base predictor, in order to isolate and highlight the contribution of spatio-temporal contextual embeddings generated by the T-STAR framework.

Table \ref{tab:accuracy_abnormal} presents the predictive accuracy of TST-based models using three different contextual input structures: \textit{Baseline}, \textit{Environment, Facility and PT}, and \textit{T-STAR}, evaluated separately over normal and abnormal demand periods. The results reveal a clear difference in model performance between the two scenarios, where prediction errors are consistently higher across all metrics during abnormal events. This highlights the inherent difficulty in forecasting rare demand fluctuations, likely due to their unpredictable nature and limited representation in historical data.
Among the models, the Transformer-based T-STAR consistently demonstrates superior predictive accuracy, particularly in uncertainty quantification. 
As expected, the TST model with \textit{Environment, Facility, and PT} inputs outperforms the simpler baseline in accuracy. This improvement is largely driven by the inclusion of facility-related features that capture censored demand effects, and real-time metro passenger flow data that reflect delays or early arrivals. Together, these factors help account for anomalies in short-term bike-sharing demand caused by unfulfilled trips and intermodal connections between metro and bike usage.
Its significantly lower MIS scores during abnormal periods suggest that the T-STAR framework enhances the model’s ability to capture and represent uncertainty, making it more reliable under rare and extreme conditions. This is enabled by the inclusion of intermediate demand variation features, which reveal unexpected sequential fluctuations in recent demand and help reduce noise during training.
These findings suggest that the Transformer-based T-STAR model is an advantageous option for supporting real-time operations in shared micro-mobility services, where anticipating and effectively managing extreme scenarios is critical in ensuring service reliability.


\begin{table}[h!]
\centering
\caption{Forecasting accuracy of TST-based models with different inputs under normal and abnormal pickup demand scenarios. \textcolor{black}{Metric values are averaged over five runs with different random seeds for each predictive model.}}
\resizebox{0.75 \textwidth}{!}{%
\begin{tabular}{l>{\color{black}}c>{\color{black}}c>{\color{black}}c>{\color{black}}c>{\color{black}}c>{\color{black}}c>{\color{black}}c>{\color{black}}c}
\hline
\textbf{Predictive Model} & \multicolumn{4}{c}{\textbf{Abnormal}} & \multicolumn{4}{c}{\textbf{Normal}} \\
\cmidrule(lr){2-5} \cmidrule(lr){6-9}
& \textbf{MAE} & \textbf{RMSE} & \textbf{MCRPS} & \textbf{MIS} 
& \textbf{MAE} & \textbf{RMSE} & \textbf{MCRPS} & \textbf{MIS} \\
\hline
\textit{Simple TST} & 2.210 & 2.497 & 1.997 & 56.662 & 0.170 & 0.440 & 0.160 & 3.264 \\
\textit{TST (\textit{Env., Facility and PT})} 
& 2.149 & 2.424 & 1.879 & 46.784 &
0.138 & 0.412 & 0.128 & 1.758 \\
\textit{Transformer-Based T-STAR} & 
\textbf{2.139} & \textbf{2.408} & \textbf{1.763} & \textbf{31.755} & \textbf{0.133} & \textbf{0.402} & \textbf{0.114} & \textbf{1.599} \\
\hline
\multicolumn{7}{l}{\makecell[l]{The best-performing values for each metric are highlighted in bold in the table.}} \\
\end{tabular}%
}
\label{tab:accuracy_abnormal}
\end{table}

\subsection{Discussion}
\label{subsect:discussion}
Building on the experimental results, this section reflects on the advantages and practical implications of the proposed Transformer-based T-STAR framework for short-term demand forecasting in shared micro-mobility systems. We discuss three key aspects that contribute to model’s effectiveness: the benefit of integrating diverse multivariate contextual inputs, the architectural advantage of T-STAR’s two-stage design in infusing multivariate inputs, and the model's generalization capability for its zero-shot forecasting application on unseen stations. Together, these elements highlight the practical strengths of the proposed predictive framework and its potential for real-world deployment for shared micro-mobility services.

\subsubsection{Analysis of Contextual Feature Contributions}
\label{subsubsect:feature_importance}
To quantitatively assess the benefits of progressively incorporating contextual information, we apply all contextual input strategies introduced in section \ref{subsubsect:contextual_input}, using the TST as the basic predictor. As shown in Table \ref{tab:contextual_inputs_results}, adding contextual features leads to steady improvements across all evaluation metrics (MAE, RMSE, MCRPS, MIS). 


\begin{table}[h!]
\centering
\caption{Short-term pickup demand forecasting performance of time-series Transformer-based models using different contextual input configurations in the Washington D.C. case study. \textcolor{black}{Metric values are averaged over five runs with different random seeds for each predictive model.} The first value in parentheses indicates the standard deviation across stations, and the second value indicates the standard deviation across time intervals (aggregated across stations at each interval).}
\resizebox{\textwidth}{!}{%
\begin{tabular}{cc>{\color{black}}c>{\color{black}}c>{\color{black}}c>{\color{black}}c}
\hline
\textbf{Basic Predictor} & \textbf{Input Types} & \textbf{MAE} & \textbf{RMSE} & \textbf{MCRPS} & \textbf{MIS} \\
\hline
TST & \textit{Baseline} & 0.188 (0.123; 0.176) & 0.501 (0.197; 0.310) & 0.177 (0.109; 0.156) & 3.753 (1.276; 2.750) \\
TST & \textit{Global Environment} & 0.152 (0.092; 0.151) & 0.469 (0.163; 0.277) & 0.131 (0.063; 0.108) & 2.248 (0.536; 0.984) \\
TST & \textit{Env., Facility and PT} & 
0.156 (0.091; 0.151) & 0.472 (0.163; 0.276) & 0.142 (0.060; 0.107) & 2.009 (0.476; 0.910) \\
TST & \textit{T-STAR} & \textbf{0.151 (0.089; 0.148)} & \textbf{0.464 (0.156; 0.271)} & \textbf{0.128 (0.062; 0.105)} & \textbf{1.864 (0.508; 0.925)} \\
\hline
\multicolumn{6}{l}{The best-performing values for each metric are highlighted in bold in the table.} \\
\end{tabular}%
}
\label{tab:contextual_inputs_results}
\end{table}

Notable gains are obtained when dynamic weather and holiday information is introduced. Moving from the simple baseline to the global environment Transformer predictor yields a consistent reduction across all error metrics. Incorporating facility and public transport signals (i.e., station capacity attributes and real-time metro flow deviations) further improves interval quality, reflected by a lower MIS, although MAE, RMSE, and MCRPS increase slightly. Across all variants, the T-STAR framework provides the best overall performance and the most stable results, as indicated by consistently low standard deviations across stations and time intervals. It achieves the lowest MAE (0.151), RMSE (0.466), and MIS (2.196), and ties for the best MCRPS (0.126). Compared to one-stage Transformer with the same set of information, the contextual T-STAR structure enhances predictive capability and model robustness, as evident by higher forecasting accuracy and smaller standard deviations. Results suggest that performance gains are driven not only by richer contextual inputs but also by T-STAR’s temporally aware design. Its distinctive two-stage structure effectively captures demand variation signals and enhances model’s adaptability to recent short-term fluctuations.


Our findings highlight the added value in integrating diverse multivariate contextual inputs in short-term demand forecasting for bike-sharing systems. As additional sources of contextual information, ranging from environmental, infrastructural, intermodal, to variational signals introduced by T-STAR, are integrated step by step, we observe consistent improvements in both predictive accuracy and uncertainty estimation. Our analysis underscores the importance of leveraging diverse signals to capture the complex demand dynamics inherent to shared micro-mobility systems.

\subsubsection{Performance Enhancement via Contextual Two-Stage Framework}
\label{subsubsect:two-stage_generalizability}

Next, we assess the added value of T-STAR’s two-stage architecture for processing multivariate contextual inputs across different predictive models. Specifically, we compare the short-term pickup demand forecasting performance of models using two contextual input strategies, one with standard \textit{Env., Facility and PT} input strategy and the other with the T-STAR framework. Both strategies use the same set of external contextual features, where the only difference lies in the structural design, as T-STAR requires no additional contextual inputs. By evaluating multiple base predictors under these two configurations, we isolate the impact of the T-STAR architecture on predictive accuracy. 


\begin{table}[h!]
\centering
\caption{Comparison of short-term pickup forecasting performance between standard (Environment and Facility) and two-stage (T-STAR) contextual input strategies across different base predictors (XGBoost, TFT, TST). \textcolor{black}{Metric values are averaged over five runs with different random seeds for each predictive model.} All models use the same set of external contextual features. The key difference lies in T-STAR’s unique two-stage architecture for processing these inputs. The first value in parentheses indicates the standard deviation across stations, and the second value indicates the standard deviation across time intervals (aggregated across stations at each interval).}
\resizebox{\textwidth}{!}{%
\begin{tabular}{cc>{\color{black}}l>{\color{black}}l>{\color{black}}l>{\color{black}}l}
\hline
\textbf{Basic Predictor} & \textbf{Input Types} & \textbf{MAE} & \textbf{RMSE} & \textbf{MCRPS} & \textbf{MIS} \\
\hline
XGBoost & \textit{Env., Facility and PT} & 0.334 (0.068; 0.120) & 0.459 (0.133; 0.205) & -- & -- \\
XGBoost & \textit{T-STAR} & 0.320\textsuperscript{\textdagger} (0.073; 0.120) & \textbf{0.450\textsuperscript{\textdagger} (0.133; 0.206)} & -- & -- \\
\hline
\textcolor{black}{STGCN} & \textit{\textcolor{black}{Env., Facilty and PT}}  &
0.422 (0.049; 0.332) & 0.713 (0.071; 0.342) & 0.399 (0.029; 0.200) & 7.465 (0.446; 1.277) \\
\textcolor{black}{STGCN} & \textit{\textcolor{black}{T-STAR}} & 0.381\textsuperscript{\textdagger} (0.064; 0.363) & 0.701\textsuperscript{\textdagger} (0.088; 0.377) &
0.372\textsuperscript{\textdagger} (0.044; 0.221) &
7.014\textsuperscript{\textdagger}  (0.726; 1.416) \\
\hline
TFT & \textit{Env., Facility and PT} & 0.156 (0.095; 0.155) & 0.473 (0.171; 0.281) & 0.134 (0.067; 0.115) & 2.123 (0.653; 1.142) \\
TFT & \textit{T-STAR} & 0.154\textsuperscript{\textdagger} (0.095; 0.154) & 0.470\textsuperscript{\textdagger} (0.176; 0.286) & 0.132\textsuperscript{\textdagger} (0.068; 0.114) & 2.060\textsuperscript{\textdagger} (0.635; 1.157) \\
\hline
TST & \textit{Env., Facility and PT} & 0.156 (0.091; 0.151) & 0.472 (0.163; 0.276) & 0.142 (0.060; 0.107) & 2.009 (0.476; 0.910) \\
TST & \textit{T-STAR} & \textbf{0.151\textsuperscript{\textdagger} (0.089; 0.148)} & 0.464\textsuperscript{\textdagger} (0.156; 0.271) & \textbf{0.128\textsuperscript{\textdagger} (0.062; 0.105)} & \textbf{1.864\textsuperscript{\textdagger} (0.508; 0.925)}\\
\hline
\multicolumn{6}{l}{(1) For each basic predictor, the best-performing value for each metric is marked with \textsuperscript{\textdagger} to indicate the best model.}\\
\multicolumn{6}{l}{(2) Across all models, the best-performing values for each metric are highlighted in bold in the table.}
\end{tabular}%
}
\label{tab:performance_enhance}
\end{table}

Results in Table \ref{tab:performance_enhance} show that the two-stage T-STAR framework consistently outperforms the standard one-stage contextual fusion approach across all \textcolor{black}{four} base models: XGBoost, \textcolor{black}{STGCN}, TFT, and TST. 
This improvement is consistent across all key metrics and is also reflected in lower spatial and temporal variability, indicating enhanced point prediction accuracy and more reliable probabilistic forecasts.
In addition, the performance boost is not limited to Transformer-based models. The decision-tree–based XGBoost \textcolor{black}{and the GNN-based STGCN} also benefit from the T-STAR framework. This suggests that the gains are not model-specific, but stem from the architectural decomposition and the use of intermediate variational representations, which can enhance learning even for non-parametric predictors. \textcolor{black}{In particular, because STGCN is sensitive to the high noise-to-signal ratio in 15-minute, station-level demand forecasting, the two-stage T-STAR design, by reducing input dimensionality and progressively revealing salient demand structure, can limit noise propagation and improve predictive performance.}
Among all configurations, the T-STAR-enhanced Transformer model delivers the best overall performance. 


The findings from this analysis reinforce the value of combining temporal attention mechanisms with structured contextual processing. By comparing two contextual fusion strategies using the same input data, we show that performance gains can be achieved without adding new features. These results highlight that the strategic processing of contextual information is a crucial consideration, next to the selection of contextual inputs. They also underscore the advantages of the two-stage T-STAR architecture, which effectively integrates diverse long-term and short-term signals to capture the complex and dynamic nature of shared micro-mobility demand.

\subsubsection{Model Generalizability and Capability for Zero-Shot Forecasting}
\label{subsubsect:model_generalization}

In rapidly expanding micro-mobility systems, new service points are frequently introduced, both within existing cities as well as in entirely new markets. During these early launch phases, providers are under pressure to deliver reliable service from the very beginning. This is especially true in competitive environments or when the brand already carries strong market expectations. To meet these expectations, operators benefit from forecasting models that require no station-specific historical data yet still provide accurate predictions. A model with zero-shot forecasting capability offers this flexibility and can be deployed immediately at new locations. Moreover, it supports pre-launch scenario testing and simulation, helping providers make informed operational decisions even before a station becomes active. Generalizable, zero-shot models are therefore valuable tools for scalable and data-efficient service planning. Recent research has shown that Transformer-based models exhibit strong zero-shot capabilities across a range of forecasting applications \citep{dooley2023forecastpfn}.

\begin{figure}[h!]
    \centering
    \includegraphics[width=0.7\linewidth]{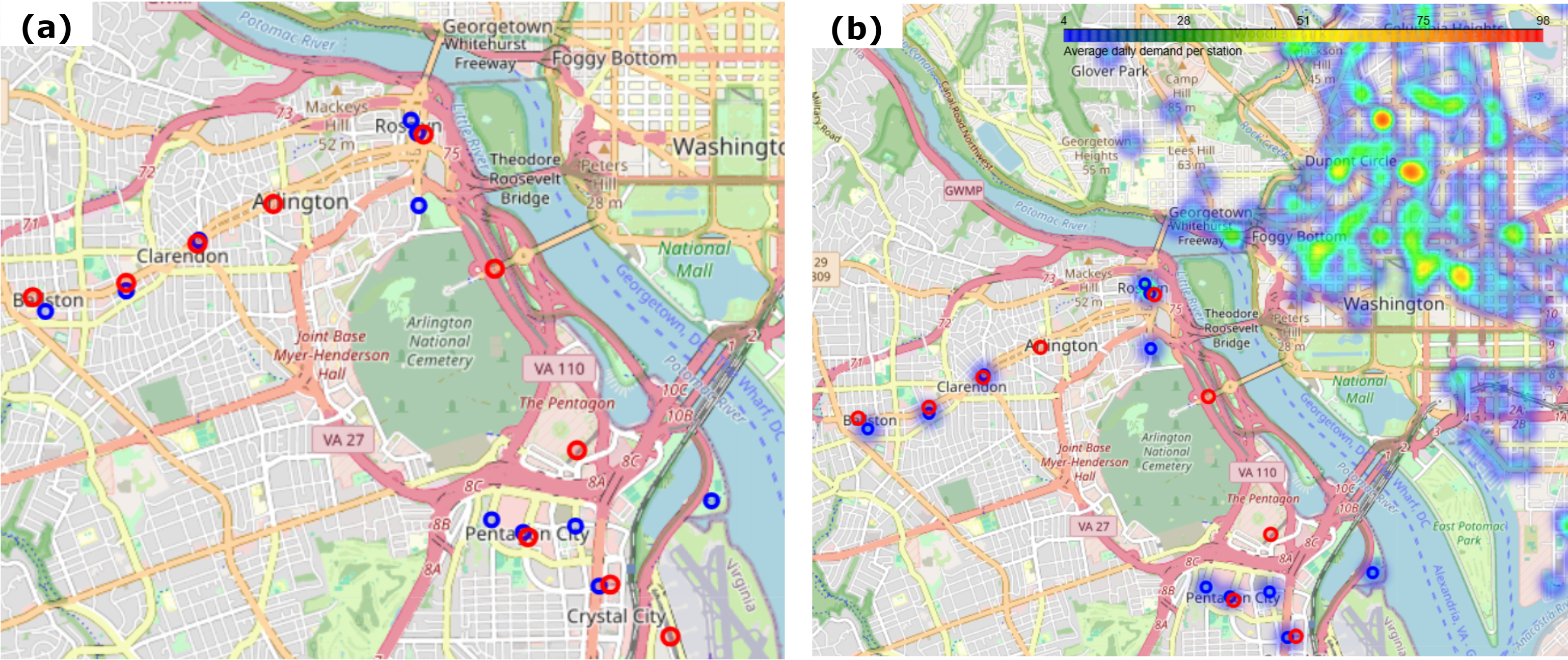}
    \caption{Spatial overview and demand comparison for zero-shot forecasting stations in the Arlington service area. Sub-figure (a) Locations of bike-sharing stations (blue circles) and metro stations (red circles) in Arlington, the unseen service area for the zero-shot forecasting experiment. Sub-figure (b) A daily average pickup demand heatmap that shows demand at the Arlington stations is generally lower than in most previously observed areas in Washington D.C.}
    \label{fig:new_station_map}
\end{figure}

To evaluate the zero-shot forecasting capability of the proposed framework, we simulate the scenario of service expansion into a new geographic area. The original model was trained on 235 bike-sharing stations located in the Washington D.C. region, all situated to the northeast of the Potomac River. However, both the WMATA metro and Capital Bikeshare services also operate beyond the D.C. boundary, including the Arlington region to the southwest of the river which lies within the state of Virginia. In this experiment, we treat Arlington as a newly added service area and select 11 bike-sharing stations located there as the test set. As shown in Figure \ref{fig:new_station_map}, many of these stations are situated near WMATA metro stations, making the setting realistic and operationally relevant.

The experiment deploys the pre-trained predictors trained solely on data from the 235 D.C. stations. No retraining or fine-tuning is performed for the Arlington stations. Instead, these predictors are applied directly, relying only on the station information and real-time inputs specified in the predictive framework. 
In this zero-shot setting, we assume no historical demand data is available for the 11 new stations, so forecasts must be generated without station-specific demand history.
For station-level embeddings, we assign each new station the mean embedding learned across the training stations, representing a typical demand characteristics in the D.C. network.
Therefore, this setup emulates a practical deployment scenario in which an operator expands into a new area and needs immediate forecasting capability without historical data.

To assess model’s generalization performance, we evaluate forecasts for the same testing period as previous experiments, focusing on the 11 Arlington stations.
We benchmark the zero-shot Transformer-based T-STAR model against \textit{Historical Average}, \textit{Myopic}, \textit{XGBoost}, \textit{DeepAR}, and \textit{TFT}. Historical Average predictor estimates are based on the historical demand pattern of the new stations from the training period. 

STGCN and STAEformer were excluded from zero-shot evaluations as they are inherently transductive. STGCN requires a predefined adjacency matrix tied to specific graph topologies, while STAEformer relies on node-specific learnable embeddings. These architectures lack the inductive capability to map static features of unseen stations to predictive representations without retraining. Unlike the other contextual predictors, they cannot effectively generalize to new locations where specific node IDs or spatial structures were not encountered during training.

Foundation time series models are large-scale predictors pretrained on massive, cross-domain datasets to capture universal temporal dependencies, allowing them to be transferred to downstream forecasting tasks without task-specific training \citep{bommasani2021opportunities}. In this study, we also include \textit{Lag-Llama}, a decoder-only transformer-based foundation model, as a competitive zero-shot benchmark \citep{rasul2023lag}. For our application, Lag-Llama utilizes the eight most recent demand observations per station as its sole input, aligning with the historical input window used by our other zero-shot predictors.

The zero-shot forecasting results are summarized in Table \ref{tab:zero-shot}. Overall, contextual models with an explicit temporal learning component (DeepAR, TFT, and T-STAR) transfer substantially better to unseen stations than naive or context-only baselines (Historical Average, Myopic, and zero-shot XGBoost). 
Interestingly, these pretrained temporal-contextual predictors also consistently outperform Lag-Llama across all four metrics. This suggests that while foundation models effectively capture universal temporal dependencies, the integration of domain-specific contextual knowledge remains vital for urban micro-mobility systems. Specifically, relying exclusively on historical lags, as Lag-Llama does, overlooks the distinct spatial and environmental profiles of new stations.

 Among the temporal deep models, performance differences are small. T-STAR achieves the best point accuracy (MAE and RMSE). DeepAR attains the lowest MCRPS and MIS, while T-STAR delivers comparable probabilistic accuracy with similar variability across stations and time intervals. Conceptually, in zero-shot deployment the stage-1 forecaster can be biased because station embeddings for new stations are not learned (we use a mean embedding). T-STAR mitigates this via stage 2, which conditions on recent deviation signals to adjust the forecasts accordingly. The same mechanism can also stabilize forecasts under abrupt demand regime shifts of stations once post-change deviations are observed. Overall, these results support T-STAR’s reliable generalization in realistic service expansion scenarios where forecasting is required immediately without station-specific historical demand. 

\begin{table}[h!]
\centering
\caption{Zero-shot per 15min pickup demand forecasting performance for 11 newly added Arlington stations using the pre-trained T-STAR model, compared with Historical Average and Myopic baselines. Metric values are averaged over five runs with different random seeds for each predictive model (except the Historical Average, Myopic approaches, and the pretrained zero-shot Lag-llama predictor).} The first value in parentheses indicates the standard deviation across stations, and the second value indicates the standard deviation across time intervals (aggregated across stations at each interval).
\resizebox{\textwidth}{!}{%
\begin{tabular}{cccccc}
\hline
\textbf{Predictive Model} & \textbf{Input Types} &\textbf{MAE} & \textbf{RMSE} & \textbf{MCRPS} & \textbf{MIS} \\
\hline
Historical Average & \textit{Baseline} &0.187 (0.032; 0.159) & 0.320 (0.064; 0.215) & -- & -- \\
Myopic & \textit{\makecell{Last demand observa-\\tion of each station}} & 0.105 (0.025; 0.150) & 0.379 (0.066; 0.300) & -- & -- \\
Zero-Shot XGBoost & \textit{Env., Facility and PT} & 0.270 (0.014; 0.083) & 0.332 (0.037; 0.148) & -- & -- \\
Zero-Shot Lag-Llama & \textit{\makecell{Previous demand observa-\\tions of each station}} & 0.075 (0.020; 0.119) & 0.307 (0.055; 0.255) & 0.069 (0.018; 0.110) & 2.357 (0.544; 4.038) \\ 
Zero-Shot DeepAR & \textit{Env., Facility and PT} & 0.063 (0.017; 0.108) & 0.292 (0.055; 0.249) & \textbf{0.059 (0.015; 0.094)} & \textbf{1.301 (0.213; 2.123)} \\
Zero-Shot TFT & \textit{Env., Facility and PT} & 0.062 (0.017; 0.108) & 0.291 (0.054; 0.200) & 0.065 (0.015; 0.089) & 1.487 (0.193; 1.374) \\
\textit{\makecell{Zero-Shot Transformer\\-Based T-STAR}} & \textit{T-STAR} & \textbf{0.061 (0.016; 0.108)} & \textbf{0.290 (0.052; 0.249)} & 0.060 (0.012; 0.085) & 1.367 (0.126; 1.268)\\
\hline
\multicolumn{5}{l}{\makecell[l]{The best-performing values for each metric are highlighted in bold in the table.}} \\
\end{tabular}%
}
\label{tab:zero-shot}
\end{table}

While the zero-shot forecasting results highlight the strong generalization capability of the Transformer-based T-STAR model, its flexibility extends beyond initial deployment. As new service areas begin operating and collect data, the demand patterns of newly added stations become more distinguishable. As shown in Figure \ref{fig:new_station_map}, Arlington stations exhibit relatively low daily pickup demand compared to most of D.C., resembling patterns seen among the lower-demand D.C. stations. In such cases, the model can be further refined through few-shot learning with limited data to fine-tune station-level embeddings and improve forecast accuracy.

Beyond zero-shot forecasting, the generalizability of Transformer-based models offers key advantages for long-term operations. Urban shared mobility markets are highly dynamic, with demand patterns continuously shifting due to changes in public transit, new points of interest, and the emergence of competing services. Transformer-based T-STAR architecture is deployment-ready to these evolving conditions thanks to their ability to support online learning and incremental model updates. 
These updates are both efficient, as they reuse prior training knowledge, and adaptive, since they require only a small amount of new data. This enables the model to respond not only to newly added service areas, but also to shifting demand patterns of current operational areas.



\section{Conclusions}
\label{chapter:conclusion}

This study addresses the challenge of reliable and accurate short-term forecasting of station-level demand in dock-based bike-sharing systems, particularly under sparse, zero-inflated, and highly variable time series. To tackle this problem, we introduced a Two-stage Spatial and Temporal Adaptive contextual Representation (T-STAR) framework that combines multivariate contextual inputs with a structured, sequential forecasting approach. By decomposing the task into two stages, T-STAR captures both long-term demand patterns and short-term fluctuations, through efficient process of coarse-grained and high-frequency contextual signals. Built on a Transformer architecture, the global T-STAR predictor enables real-time forecasting across large-scale bike-sharing networks, while adapting effectively to varying urban demand conditions. 
To evaluate the proposed framework, we conducted 15-minute demand forecasting experiments using real-world data from Washington D.C.’s Capital Bikeshare system. We assessed both deterministic and probabilistic forecasting across different demand intensities, time periods, and station locations. The Transformer-based T-STAR model consistently delivered robust and accurate predictions, particularly at high-demand stations and under irregular demand conditions.

From a managerial perspective, this study highlights the practical value of adopting a hierarchical forecasting framework to effectively process the diverse contextual signals inherent in shared micro-mobility systems. By leveraging long-term demand trends, the T-STAR model enhances the robustness of short-term forecasts through reduced training noise. Incorporating rich external context—such as weather, infrastructure, and real-time intermodal dynamics—further improves forecast accuracy and resilience. These findings emphasize the importance of strategic data sharing and collaboration between public transport and shared mobility providers to strengthen multimodal integration and enhance the user experience within MaaS ecosystems.
Transformer-based T-STAR’s strong generalization capabilities also support zero-shot forecasting for new service areas and enable few-shot adaptation as data becomes available, offering crucial flexibility for fast-growing, dynamic urban mobility networks. Furthermore, integrating fine-grained, real-time forecasts into operational decision-making presents a valuable opportunity. Applications such as dynamic fleet rebalancing \citep{zhao2025research} and multimodal trip coordination via MaaS platforms \citep{zhang2025stochastic} can greatly benefit from these predictive insights, improving the efficiency and reliability of shared mobility services in evolving urban environments.

Future research could build upon this study in several key dimensions, addressing both the current data constraints and potential architectural extensions. 
While the rolling-origin and sliding-window evaluations demonstrate the robustness of our results across different temporal splits, the three-month dataset limits the assessment of long-term predictive performance under diverse seasonal related impacts, such as weather changes, holiday periods, and tourism-driven demand changes. 
Therefore, a key future research direction involves extending the T-STAR framework to multi-year datasets to capture broader cyclical patterns and seasonal shifts.
Beyond temporal coverage, future work could evolve T-STAR into a unified multi-task learning framework. Rather than modeling pickup and drop-off demand independently, a joint formulation could simultaneously forecast both demand streams and station-level fleet availability, once historical inventory observations are accessible. Such integration would better capture the coupled spatio-temporal dynamics of supply and demand. In particular, conditioning on real-time fleet distribution would enable direct estimation of station-wise net flow, providing an operationally meaningful signal for anticipating stock-outs and surpluses. A comprehensive predictor of this form could support earlier identification of rebalancing requirements and improve system reliability by enabling more informed, forward-looking control decisions.
Future work can also strengthen T-STAR by coupling its two-stage temporal design with dynamic, attention-driven GNNs for high-granularity, intermittent demand. A learnable time-varying graph inferred from demand co-movement, operational interactions, and context (weather, events, commute regimes) would enable message passing only when informative. Attention-based neighbor selection and heterophily-robust layers can reduce noise, while interaction edges capture competition/substitution to improve fluctuation modeling and calibration.
Finally, future research could explore the few-shot learning capabilities of the T-STAR framework and its generalizability across different service regions with limited historical data.
While this study focuses on zero-shot forecasting, T-STAR’s broader potential lies in its ability to adapt to new environments through lightweight fine-tuning. Investigating how the model can respond to evolving urban conditions would further demonstrate its value as a flexible and scalable tool for long-term shared micro-mobility management.

\section{Acknowledgments}
This research is supported by the project Seamless Shared Urban Mobility (SUM), which received funding from the European Union’s Horizon 2020 research and innovation program under grant agreement no. 101103646. The authors would also like to thank the Washington Metropolitan Area Transit Agency (WMATA) for their support in providing the data used in this study.

\bibliographystyle{elsarticle-harv} 
\bibliography{example}

\newpage

\appendix

\section{Negative Binomial Distribution}
\label{appendix:NB}

The Negative Binomial (NB) distribution is a discrete probability distribution used to model the number of event occurrences (e.g., bike pick-ups or drop-offs) within a fixed time interval. Unlike the Poisson distribution, which assumes equal mean and variance, the NB distribution includes a dispersion parameter that allows for greater variance. This flexibility makes it well-suited for over-dispersed count data and enables more accurate and robust probabilistic forecasting.

Assuming $k$ is the number of observed demand, $r$ is the shape parameter that reflecting the degree of demand variability, $p$ is the probability parameter relating to the demand intensity level,
the probability mass function of Negative Binomial distribution is given by,

\[
P(X = k) = \binom{k + r - 1}{k} (1 - p)^r p^k.
\]

The expected demand $\mu$, which is the mean of a NB distribution, is expressed as:

\[
\mu = \frac{r \cdot p}{1 - p}.
\]

Substituting $p$ in terms of $\mu$, we obtain:

\[
p = \frac{\mu}{r + \mu}.
\]

Thus, the probability mass function can be rewritten in terms of the expected demand $\mu$ and the shape parameter $r$, both of which are estimated by the Time Series Transformer model, as:

\[
P(X = k) = \binom{k + r - 1}{k} \left( \frac{r}{r + \mu} \right)^r \left( \frac{\mu}{r + \mu} \right)^k.
\]

The variance of Negative Binomial distribution is given by $\mu + r^{-1} \mu^2$, where $\mu$ is the mean value.

\begin{figure}[h!]
    \centering
    \includegraphics[width=0.6\linewidth]{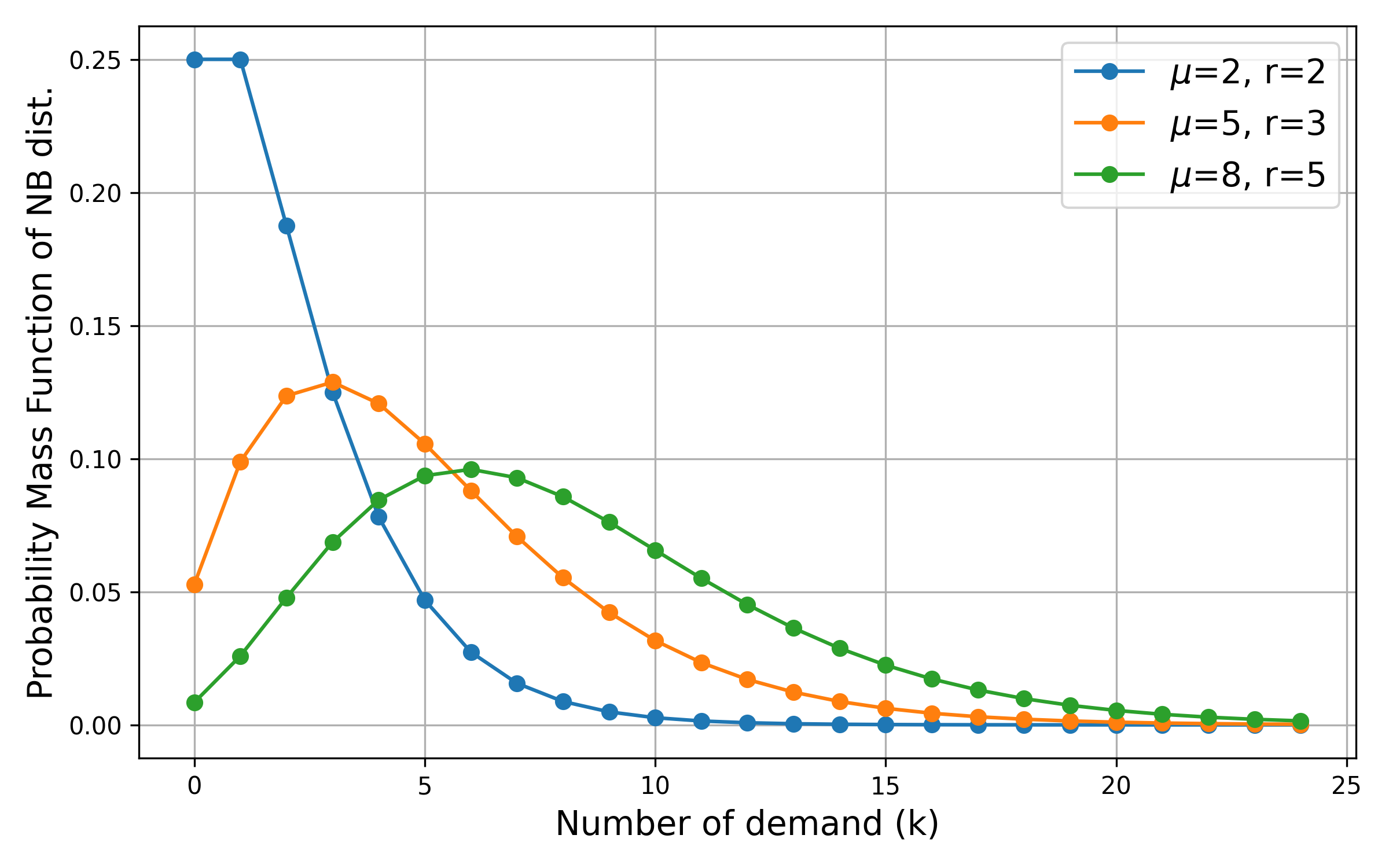}
    \caption{Negative binomial distributions for different pairs of $\mu$ and $r$.}
    \label{fig:NB_dist_example}
\end{figure}

\section{Graph Construction Strategies for STGCN and STGCN-VAE}
\label{appendix:graph_construction}

Following the framework of \cite{yu2017spatio},
we represent the bike-sharing network as a weighted graph $\mathcal{G} = (\mathcal{V}, \mathcal{E}, \mathbf{A})$, where $\mathcal{V}$ denotes the set of $N$ stations and the adjacency matrix $\mathbf{A} \in \mathbb{R}^{N \times N}$ encodes the underlying spatial dependencies. 
To ensure architectural consistency with our Transformer-based predictors, station-specific attributes are incorporated as exogenous node features in both models, enabling the learning of high-dimensional latent station embeddings.
In order to adapt STGCN for probabilistic demand forecasting, we appended a distributional output head to the final layer, allowing the model to parameterize a predictive distribution (e.g., Gaussian or Poisson). The parameters of this distribution are optimized by minimizing the NLL. In contrast, for STGCN-VAE, probabilistic density is captured through the inherent stochasticity of the Variational Autoencoder framework, where the decoder reconstructs the future demand distribution based on the sampled latent variables.

To ensure a robust evaluation of our STGCN and STGCN-VAE benchmarks, we systematically investigated three distinct strategies for constructing the static adjacency matrix $\mathbf{A}$: (i) Location Proximity, (ii) Trip-Flow Connectivity, and (iii) Demand Correlation. These strategies allow the models to capture dependencies driven by geographic distance, physical movement patterns, and functional usage similarity, respectively.

\subsection*{Location proximity graph}
In the first graph construction strategy, the static adjacency matrix is calculated based on geographic proximity. Specifically, we employ an Inverse Distance Weighting (IDW) scheme with a local threshold (1000m) to define the edge weights $A_{ij}$ between stations $i$ and $j$:

\begin{equation}
A_{ij} = 
\begin{cases} 
\frac{1}{d_{ij}}, & \text{if } 0 < d_{ij} \leq 1000\text{m} \\ 
0, & \text{otherwise}
\end{cases}
\end{equation}

where $d_{ij}$ denotes the pairwise geodesic distance in meters. This construction, adapted from \cite{liang2022bike}, ensures the model captures strong local spatial correlations while maintaining a sparse adjacency matrix to mitigate computational overhead.

\subsection*{Trip-flow based graph}

The second strategy constructs the adjacency matrix based on the observed origin-destination (OD) trip flows, capturing the latent connectivity induced by the physical movement of the bike fleet. Let $F_{ij}$ denote the total number of trips originating at station $i$ and terminating at station $j$ during the training period. Since STGCN typically operates on undirected graphs, we define a symmetric connectivity weight $W_{ij}$ by aggregating bidirectional flows:
\begin{equation}
W_{ij} = F_{ij} + F_{ji}
\end{equation}
This matrix $W$ reflects the global intensity of direct interactions between station pairs. 
To ensure a sparse and computationally efficient graph structure, we apply a top-$K$ selection rule. The final adjacency matrix $\mathbf{A}$ is constructed by retaining only the strongest connections for each node:
\begin{equation}
A_{ij} =
\begin{cases}
W_{ij}, & \text{if } W_{ij} \in \text{top-}K \text{ connections for station i or j}\\
0, & \text{otherwise}\end{cases}\end{equation}
where the $A_{ij}$ is non-zero when the trip flow connection is among the top-$K$ for either station $i$ or $j$. Parameter $k$ is selected to be five in our case study.

\subsection*{Demand correlation graph}

In the final strategy, spatial dependencies are derived from the statistical correlations between station-level demand time series. This approach identifies functional similarities between stations that may be geographically distant but exhibit synchronized usage patterns. For each station pair $(i, j)$, we compute the Pearson correlation coefficient $\rho_{i,j}$ based on their historical demand profiles, following the design of \cite{ke2021predicting}. To focus on meaningful positive dependencies and ensure network sparsity, we define the adjacency matrix $\mathbf{A}$ using a percentile-based threshold $\tau$:
$$A_{i,j} =\begin{cases}\rho_{i,j} & \text{if } \rho_{i,j} \geq \tau \text{ and } i \neq j \\ 0 & \text{otherwise}\end{cases}$$
where $\tau$ is set to the $95^{th}$ percentile of all positive off-diagonal correlation values. By retaining only the top 5\% of correlations, the resulting graph effectively isolates the most significant functional relationships within the network.

\subsection*{Comparative Performance}
The comparative performance of STGCN and STGCN-VAE across the three graph construction strategies is summarized in Tables \ref{tab:pickup_graph_accuracy} and \ref{tab:dropoff_graph_accuracy}. To ensure a fair and robust comparison, we conducted extensive hyperparameter tuning for each model-graph configuration, and all reported metrics represent the average of five independent runs with different random seeds.
Our findings indicate that for the standard STGCN, the Location Proximity graph consistently yielded the most stable and accurate results, outperforming both trip-flow and demand-correlation configurations. Conversely, the probabilistic STGCN-VAE exhibited greater resilience to the underlying graph topology, with all three strategies achieving significantly improved and comparable accuracy. Notably, the combination with a Demand Correlation graph provided the best performance for STGCN-VAE in drop-off forecasting, while the one with Trip-Flow graph was slightly superior for pickups. For the final results reported in Section \ref{chapter:results_and_discussion}, we have selected the best-performing graph construction choice for both STGCN and STGCN-VAE to reflect the best performance of these models.

\begin{table*}[h!]
\centering
\caption{Pickup forecasting performance of STGCN and STGCN-VAE across different graph construction strategies.}
\label{tab:pickup_graph_accuracy}
\resizebox{\textwidth}{!}{
\begin{tabular}{llcccc}
\hline
\hline
\multirow{2}{*}{Model} & \multirow{2}{*}{Graph Type} 
& \multicolumn{4}{c}{Pickup} \\
\cline{3-6}
& & MAE & RMSE & CRPS & IS \\
\hline

STGCN & Location Proximity Graph
& \textbf{0.422}  (0.049, 0.332) 
& \textbf{0.713}  (0.071, 0.342) 
& \textbf{0.399}  (0.029, 0.200) 
& \textbf{7.465}  (0.446, 1.142) \\

STGCN & Trip Flow Connection Graph
& 0.650  (0.066, 0.514) 
& 0.986  (0.084, 0.502) 
& 0.577  (0.052, 0.307) 
& 9.889  (0.938, 2.249) \\

STGCN & Demand Correlation Graph
& 0.545  (0.117, 0.477) 
& 0.897  (0.122, 0.475) 
& 0.495  (0.075, 0.297) 
& 8.416  (0.735, 2.139) \\

\hline

STGCN-VAE & Location Proximity Graph
& 0.405  (0.057, 0.087) 
& 0.501  (0.124, 0.175) 
& 0.402  (0.057, 0.087) 
& 15.832  (2.282, 3.456)\\

STGCN-VAE & Trip Flow Correlation Graph
& \textbf{0.404}  (0.057, 0.087) 
& \textbf{0.500}  (0.124, 0.175) 
& \textbf{0.398}  (0.057, 0.087) 
& \textbf{15.419}  (2.286, 3.463) \\

STGCN-VAE & Demand Correlation Graph 
& 0.410  (0.056, 0.085) 
& 0.503  (0.122, 0.173) 
& 0.402  (0.056, 0.085) 
& 15.398  (2.246, 3.394) \\

\hline
\hline
\end{tabular}
}
\end{table*}

\begin{table*}[h!]
\centering
\caption{Drop-off forecasting performance of STGCN and STGCN-VAE across different graph construction strategies.}
\label{tab:dropoff_graph_accuracy}
\resizebox{\textwidth}{!}{
\begin{tabular}{llcccc}
\hline
\hline
\multirow{2}{*}{Model} & \multirow{2}{*}{Graph Type} 
& \multicolumn{4}{c}{Drop-off} \\
\cline{3-6}
& & MAE & RMSE & CRPS & IS \\
\hline

STGCN & Location Proximity Graph
& \textbf{0.602}  (0.066, 0.479) 
& \textbf{0.917}  (0.069, 0.461) 
& \textbf{0.535}  (0.036, 0.283) 
& \textbf{8.969}  (0.201, 1.946) \\

STGCN & Trip Flow Connection Graph
& 0.792  (0.026, 0.521) 
& 1.107  (0.039, 0.493) 
& 0.695  (0.024, 0.307) 
& 11.388  (0.753, 3.522) \\

STGCN & Demand Correlation Graph
& 0.881  (0.163, 0.780) 
& 1.340  (0.181, 0.757) 
& 0.725  (0.104, 0.491) 
& 10.418  (0.918, 3.353) \\

\hline

STGCN-VAE & Location Proximity Graph
& 0.405  (0.058, 0.089) 
& 0.503  (0.128, 0.184) 
& 0.404  (0.058, 0.089) 
& 16.046  (2.306, 3.557) \\

STGCN-VAE & Trip Flow Correlation Graph
& 0.411  (0.057, 0.088) 
& 0.506  (0.127, 0.181) 
& 0.405  (0.057, 0.088) 
& 15.739  (2.272, 3.498) \\

STGCN-VAE & Demand Correlation Graph
& \textbf{0.395}  (0.059, 0.092) 
& \textbf{0.498}  (0.132, 0.188) 
& \textbf{0.390}  (0.059, 0.092) 
& \textbf{15.171}  (2.369, 3.662) \\

\hline
\hline
\end{tabular}
}
\end{table*}

\section{Hyperparameter Tuning for Predictive Models}
\label{appendix:hyperparameter}

This section presents the hyperparameter ranges explored during model tuning, along with the final selected values for each predictive model. All models use a consistent look-back window of 30 time steps, meaning the previous 30 observations from the target time series are used as input for forecasting. The hyperparameter tuning procedures follow the methods described in section \ref{subsect:base_predictors}. Below, we outline the specific hyperparameters tuned during the experiments and the corresponding values chosen based on validation performance.

\subsection*{Time Series Transformer}
Both stages of the Transformer-based T-STAR model, as well as the benchmark Simple TST model, were implemented using the Hugging Face library. 
Three hyperparameters are selected for fine-tuning: \textit{number of encoder layers ($N_E$)} determines the number of encoder blocks included in TST, \textit{hidden size} sets the hidden layer dimension for the self-attention layer in a Transformer model, \textit{dropout rate} is a regularization hyperparameter applied to avoid overfittting, which controls the ratio of neurons being randomly deactivated during a forward pass. The \textit{learning rate} controls how quickly the model updates its weights during training process.
The value ranges explored for key architectural hyperparameters are summarized in Table \ref{tab:transformer_hyperparams}, along with the final selected values for each model.

\begin{table}[h!]
\centering
\small 
\caption{\textcolor{black}{Tuning ranges and selected hyperparameter values for time series Transformer-based models.}}
\begin{tabularx}{\textwidth}{l c *{6}{C}}
\toprule
& & \multicolumn{3}{c}{\textbf{Pickup}} & \multicolumn{3}{c}{\textbf{Dropoff}}\\
\cmidrule(lr){3-5} \cmidrule(lr){6-8}
\textbf{Hyperparameter} & \makecell{\textbf{Value} \\ \textbf{Range}} & \textbf{T-STAR: Stage 1} & \textbf{T-STAR: Stage 2} & \textbf{Simple TST} & \textbf{T-STAR: Stage 1} & \textbf{T-STAR: Stage 2} & \textbf{Simple TST}\\
\midrule
$N_E$ & [1, 2, 3]& 1 & 3 & 2 & 2 & 2 & 2 \\
\addlinespace
hidden size & [16, 32, 64] & 64 & 16 & 64 & 64 & 64 & 64 \\
\addlinespace
dropout rate & [0.1, 0.2, 0.3] & 0.1 & 0.1 & 0.3 & 0.2 & 0.2 & 0.2 \\
\addlinespace
learning rate & [$5 \times 10^{-5}$, 0.01] & $6 \times 10^{-4}$ & - & $2.8 \times 10^{-4}$ & $1.5 \times 10^{-4}$ &  & $9.5 \times 10^{-5}$\\
\bottomrule
\multicolumn{8}{l}{\footnotesize $N_E$: number of encoder layers.}
\end{tabularx}
\label{tab:transformer_hyperparams}
\end{table}

\subsection*{XGBoost}
Five key hyperparameters were fine-tuned for the XGBoost model. The \textit{maximum depth} parameter controls the maximum depth of each decision tree. While deeper trees can capture more complex patterns, they also increase the risk of overfitting. The \textit{number of estimators} defines the total number of trees used in the boosting process. To introduce randomness and prevent overfitting, the \textit{subsample} parameter specifies the fraction of the training data used to grow each tree. Similarly, \textit{colsample bytree} controls the proportion of input features randomly selected for each tree, promoting model diversity and robustness. The \textit{learning rate} determines how much each tree contributes to the final prediction. A smaller learning rate results in more stable learning and better generalization but requires more trees, whereas a larger learning rate speeds up training at the cost of a higher risk of overfitting. 

Table \ref{tab:xgb_hyperparams} presents the tuning ranges of the hyperparameters for the XGBoost predictor, along with the values selected through the fine-tuning process.

\begin{table}[h!]
\centering
\small 
\caption{Hyperparameter tuning ranges and selected values for XGBoost model.}
\begin{tabularx}{0.6 \textwidth}{l c c c} 
\toprule
\textbf{Hyperparameter} & \textbf{Value Range} & \textbf{Pickup} & \textbf{Dropoff}\\
\midrule
maximum depth       & [10, 25, 30, 45, 60]      & 30 & 30\\
\addlinespace
number of estimators & [50, 100, 150, 200]    &  150 & 150\\
\addlinespace
subsample           & [0.3, 0.5, 0.7, 1.0]    & 1.0 & 1.0 \\
\addlinespace
colsample bytree    & [0.3, 0.5, 0.7, 1.0]    & 0.7 & 0.7\\
\addlinespace
learning rate       & [$5 \times 10^{-5}$, 0.01]   & $8 \times 10^{-3}$ & $9.6 \times 10^{-3}$\\
\bottomrule
\end{tabularx}
\label{tab:xgb_hyperparams}
\end{table}

\subsection*{DeepAR}
The DeepAR model implemented via GluonTS was fine-tuned using three key hyperparameters. The \textit{number of layers} parameter specifies the number of RNN layers in the model, determining the depth of temporal learning. The \textit{hidden size} (also referred to as number of cells) defines the dimensionality of the hidden state in RNN layer. 
Table \ref{tab:deepar_hyperparams} summarizes the value ranges used during tuning and the selected values of these hyperparameters for the DeepAR model.

\begin{table}[H]
\centering
\small 
\caption{Hyperparameter tuning ranges and selected values for DeepAR model.}
\begin{tabularx}{0.5 \textwidth}{l c c c}
\toprule
\textbf{Hyperparameter} & \textbf{Value Range} & \textbf{Pickup} & \textbf{Dropoff}\\
\midrule
number of layers    & [1, 2, 3]        & 3      & 3\\
\addlinespace
hidden size         & [32, 64, 128]   & 128     & 32 \\
\addlinespace
dropout rate        & [0.1, 0.2, 0.3] & 0.1     & 0.2 \\
\addlinespace
learning rate      & [$5 \times 10^{-5}$, 0.01] & $8 \times 10^{-5}$ & $1.3 \times 10^{-4}$ \\
\bottomrule
\end{tabularx}
\label{tab:deepar_hyperparams}
\end{table}

\subsection*{Temporal Fusion Transformer (TFT)}
For TFT predictors, we tuned the classic neural network hyperparameters \textit{learning rate}, \textit{dropout rate}, and three hyperparameters specified for the Transformer structure. 
The first structural hyperparameter determines the \textit{number of attention heads} in self-attention layer in the decoder, allowing the model to focus on different aspects of the input sequence. 
The \textit{hidden size} controls the size of hidden states for LSTM and Transformer layers. The \textit{hidden size} should be divisible by number of attention heads.
The \textit{embedding dimension} sets the size of feature embedding tensors.
Table \ref{tab:tft_hyperparams} presents the tuned hyperparameters for TFT, including their search ranges and selected values.

\begin{table}[H]
\centering
\small 
\caption{Hyperparameter tuning ranges and selected values for Temporal Fusion Transformer (TFT).}
\begin{tabularx}{0.6 \textwidth}{l c c c}
\toprule
\textbf{Hyperparameter} & \textbf{Value Range} & \textbf{Pickup} & \textbf{Dropoff}\\
\midrule
number of attention heads & \{1, 2, 3\} & 1 & 3 \\
\addlinespace
hidden size & $[4, 8, 16, 32] \times \text{heads}$ & 32 & 16 \\
\addlinespace
embedding dimension & $[4, 16, 32] \in \mathbb{N}$ & 16 & 16 \\
\addlinespace
dropout rate        & [0.1, 0.2, 0.3] & 0.1     & 0.3 \\
\addlinespace
learning rate      & [$5 \times 10^{-5}$, 0.01] & $8 \times 10^{-5}$ & $1.3 \times 10^{-4}$ \\
\bottomrule
\end{tabularx}
\label{tab:tft_hyperparams}
\end{table}

\subsection*{Spatio-Temporal Graph Convolutional Network (STGCN) and the Variational Autoencoder variant (STGCN-VAE)}

For the STGCN and STGCN-VAE models, in addition to the \textit{learning rate} and \textit{dropout rate}, we tuned the \textit{hidden channels} and the \textit{number of STGCN blocks}. The \textit{number of STGCN blocks} corresponds to the number of spatio-temporal convolution blocks stacked sequentially, and thus controls the depth of the network. Increasing this depth can improve the model’s ability to capture higher-order and longer-range spatio-temporal dependencies, but it also increases computational cost and may raise the risk of overfitting. The \textit{hidden channels} parameter specifies the dimensionality of the latent feature representation within the STConvBlock backbone, thereby controlling the capacity of the model’s internal embeddings for each node and time step.
Table \ref{tab:stgcn_hyperparams} shows the fine-tuned hyperparameters for STGCN models, 
and Table \ref{tab:stgcn_vae_hyperparams} shows those for STGCN-VAE models. Note that the number of STGCN blocks in STGCN-VAE is fixed to be two according to \cite{peng2025uncertainty}.

\begin{table}[H]
\centering
\small 
\caption{Hyperparameter tuning ranges and selected values for Spatio-Temporal Graph Convolutional Network (STGCN).}
\begin{tabularx}{0.5 \textwidth}{l c c c}
\toprule
\textbf{Hyperparameter} & \textbf{Value Range} & \textbf{Pickup} & \textbf{Dropoff}\\
\midrule
$N_B$ & [1, 2, 3]& 1 & 3 \\
\addlinespace
hidden channels & [16, 32, 64] & 16 & 16 \\
\addlinespace
dropout rate & [0.1, 0.2, 0.3] & 0.1 & 0.1\\
\addlinespace
learning rate & [$5 \times 10^{-5}$, 0.01] & $5.4 \times 10^{-5}$ & $5.6 \times 10^{-5}$ \\
\bottomrule
\multicolumn{4}{l}{\footnotesize $N_B$: number of STGCN blocks in the predictive model.}
\end{tabularx}
\label{tab:stgcn_hyperparams}
\end{table}

\begin{table}[H]
\centering
\small 
\caption{Hyperparameter tuning ranges and selected values for Spatio-Temporal Graph Convolutional Network Variational Autoencoder (STGCN-VAE).}
\begin{tabularx}{0.5 \textwidth}{l c c c}
\toprule
\textbf{Hyperparameter} & \textbf{Value Range} & \textbf{Pickup} & \textbf{Dropoff}\\
\midrule
hidden channels & [16, 32, 64] & 64 & 64 \\
\addlinespace
dropout rate & [0.1, 0.2, 0.3] & 0.2 & 0.2\\
\addlinespace
learning rate & [$5 \times 10^{-5}$, 0.01] & $9.5 \times 10^{-5}$ & $2.5 \times 10^{-4}$ \\
\bottomrule
\multicolumn{4}{l}{\footnotesize $N_B$: number of STGCN blocks in the predictive model.}
\end{tabularx}
\label{tab:stgcn_vae_hyperparams}
\end{table}

\subsection*{\textcolor{black}{Spatio-Temporal Adaptive Embedding Transformer (STAEformer)}}

\textcolor{black}{In addition to the standard \textit{learning rate} and \textit{dropout rate}, we tuned three architecture-specific hyperparameters for STAEformer. The \textit{number of self-attention layers} controls the encoder depth by specifying how many self-attention blocks are stacked in the temporal-attention module and the spatial-attention module. The \textit{spatial embedding size} determines the dimensionality of the learned static node (station) embeddings, which provide a node-identity representation shared across time. The \textit{adaptive embedding size} sets the dimensionality of the learned time-by-node adaptive embeddings, which supply an additional spatio-temporal latent representation concatenated to the input features. Table \ref{tab:STAEformer_hyperparams} presents the tuning ranges and selected values for these hyperparameters. }

\begin{table}[H]
\centering
\small 
\caption{\textcolor{black}{Hyperparameter tuning ranges and selected values for Spatio-Temporal Adaptive Embedding Transformer (STAEformer).}}
\begin{tabularx}{0.55 \textwidth}{l c c c}
\toprule
\textbf{Hyperparameter} & \textbf{Value Range} & \textbf{Pickup} & \textbf{Dropoff}\\
\midrule
$N_S$ & [1, 2, 3]& 2 & 1 \\
\addlinespace
spatial embedding size & [8, 16, 32] & 16 & 32 \\
\addlinespace
adaptive embedding size & [8, 16, 32] & 8 & 8 \\
\addlinespace
dropout rate & [0.1, 0.2, 0.3] & 0.2 & 0.3\\
\addlinespace
learning rate & [$5 \times 10^{-5}$, 0.01] & $1.5 \times 10^{-4}$ & $5 \times 10^{-4}$ \\
\bottomrule
\multicolumn{4}{l}{\footnotesize $N_S$: number of self-attention layers in the predictive model.}
\end{tabularx}
\label{tab:STAEformer_hyperparams}
\end{table}

\section{\textcolor{black}{Robustness Analysis of Transformer-Based T-STAR}}
\label{appendix:robustness}

\textcolor{black}{To evaluate the robustness of the Transformer-Based T-STAR predictor, we performed a rolling forecast origin (expanding-window) experiment and sliding window cross-validation using the Washington D.C. dataset. These evaluations are designed to test the model's sensitivity to training data volume and its stability across varying temporal contexts. To isolate the benefits of the T-STAR architecture, we benchmark its performance against a Contextual TST baseline, which incorporates the same exogenous features (Environment, Facility, and Public Transport inputs) but lacks the T-STAR structural design.}

\textcolor{black}{The rolling forecast origin experiment simulates a real-world deployment pipeline where the model is periodically retrained as new data is ingested. We initialized the training set using the first four weeks ($w_1\text{-}w_4$) and expanded the horizon incrementally to ten weeks ($w_1 \text{-} w_{10}$). To prevent data leakage, each model was evaluated on the two-week period immediately following the training boundary.} \textcolor{black}{As detailed in Table \ref{tab:rfo_robustness_pickup_dropoff}, MAE generally follows a downward trend for both models as the training history increases, confirming the expected benefit of larger data volumes. Notably, however, the T-STAR Transformer consistently outperforms the Contextual TST baseline across every fold. The performance margin is particularly significant in the initial folds ($1^{st}$ and $2^{nd}$), where training data is most constrained. This demonstrates that T-STAR Transformer is not only more accurate overall but also significantly more data-efficient, maintaining robust predictive performance even when historical observations are sparse.}

\begin{table*}[h!]
\centering
\small
\setlength{\tabcolsep}{6pt}
\caption{\textcolor{black}{Rolling forecast origin robustness results. Each MAE entry reports the mean MAE across stations. Values in parentheses are the station-wise standard deviation. $w_i$ denotes the $i^{th}$ week in the Washington D.C. case study dataset.}}
\label{tab:rfo_robustness_pickup_dropoff}
\begin{tabular}{cccccccc}
\toprule
\multirow{2}{*}{Fold} & \multirow{2}{*}{\shortstack{Train\\weeks}} & \multirow{2}{*}{\shortstack{Validation\\weeks}} &
\multicolumn{2}{c}{Pickup MAE} & \multicolumn{2}{c}{Drop-off MAE} \\
\cmidrule(lr){4-5}\cmidrule(lr){6-7}
& & & Contextual TST & T-STAR Transformer & Contextual TST &  T-STAR Transformer \\
\midrule
$1^{th}$ & $w_1$--$w_4$  & $w_5$--$w_6$ & 0.365 (0.175) & \textbf{0.326 (0.158)} & 0.361 (0.173) & \textbf{0.317 (0.152)} \\
$2^{th}$ & $w_1$--$w_6$  & $w_7$--$w_8$ & 0.237 (0.123) & \textbf{0.210 (0.113)} & 0.469 (0.263) & \textbf{0.207 (0.115)} \\
$3^{th}$ & $w_1$--$w_8$  & $w_9$--$w_{10}$ & 0.253 (0.146) & \textbf{0.238 (0.129)} & 0.253 (0.143) & \textbf{0.239 (0.133)} \\
$4^{th}$ & $w_1$--$w_{10}$ & $w_{10}$--$w_{11}$ & 0.172 (0.106) & \textbf{0.163 (0.096)} & 0.183 (0.105) & \textbf{0.163 (0.094)} \\
\bottomrule
\end{tabular}
\end{table*}

\textcolor{black}{To assess the model’s sensitivity to short-term distribution shifts and varying historical contexts, we conducted a sliding window cross-validation. This setup utilizes a fixed 8-week training duration and a 2-week validation period, advancing the window forward by one week for each subsequent fold. This approach ensures that the model is tested against different temporal segments of the demand distribution.}
\textcolor{black}{As shown in Table \ref{tab:sfwcv_robustness_pickup_dropoff}, although forecast difficulty fluctuates across folds (likely reflecting localized events or inherent seasonality in urban mobility), T-STAR Transformer delivers uniformly lower MAE and reduced station-level standard deviation compared to the Contextual TST benchmark. This consistent performance across shifting temporal origins underscores T-STAR’s superior ability to generalize across evolving demand patterns while maintaining high spatial precision. The reduction in standard deviation also indicates that T-STAR provides more reliable and equitable predictions across the entire station network.}

\begin{table*}[h!]
\centering
\small
\setlength{\tabcolsep}{6pt}
\caption{\textcolor{black}{Sliding forecast window cross validation results. Each MAE entry reports the mean MAE across stations. Values in parentheses are the station-wise standard deviation. $w_i$ denotes the $i^{th}$ week in the Washington D.C. case study dataset.}}
\label{tab:sfwcv_robustness_pickup_dropoff}
\begin{tabular}{cccccccc}
\toprule
\multirow{2}{*}{Fold} 
& \multirow{2}{*}{\shortstack{Train\\weeks}}
& \multirow{2}{*}{\shortstack{Validation\\weeks}}
& \multicolumn{2}{c}{Pickup MAE}
& \multicolumn{2}{c}{Drop-off MAE} \\
\cmidrule(lr){4-5}\cmidrule(lr){6-7}
& & & Contextual TST &  T-STAR Transformer & Contextual TST &  T-STAR Transformer \\
\midrule
$1^{st}$ & $w_1$--$w_8$  & $w_9$--$w_{10}$ & 0.253 (0.146) & \textbf{0.238 (0.129)} & 0.253 (0.143) & \textbf{0.239 (0.133)} \\
$2^{nd}$ & $w_2$--$w_9$  & $w_{10}$--$w_{11}$ & 0.273 (0.159) & \textbf{0.228 (0.130)} & 0.256 (0.134) & \textbf{0.225 (0.127)} \\
$3^{rd}$ & $w_3$--$w_{10}$ & $w_{11}$--$w_{12}$ & 0.174 (0.108) & \textbf{0.166 (0.098)} & 0.175 (0.102) & \textbf{0.169 (0.104)} \\
$4^{th}$ & $w_4$--$w_{11}$ & $w_{12}$--$w_{13}$ & 0.134 (0.078) & \textbf{0.121 (0.071)} & 0.124 (0.076) & \textbf{0.119 (0.072)} \\
\bottomrule
\end{tabular}
\end{table*}

\textcolor{black}{Across both evaluations, T-STAR Transformer consistently maintains a decisive performance edge over the Contextual TST benchmark. The simultaneous reduction in mean absolute error and station-wise variance confirms that the T-STAR architecture is not only more accurate in its point estimates but also significantly more stable under diverse operational conditions.}







\end{document}